\documentclass[preprint,5p,twocolumn,authoryear]{elsarticle}

 \usepackage[table,xcdraw]{xcolor}
\usepackage{amssymb}
\usepackage{blindtext}
\usepackage{svg}
\usepackage{graphicx}
\usepackage{epstopdf}
\usepackage[noend]{algpseudocode}
\usepackage{amsmath}
\usepackage{gensymb}
\usepackage{enumitem}
\usepackage{tabularx}
\usepackage{svg}
\usepackage{xcolor}
\setsvg{inkscape=inkscape -z -D,svgpath=images/}
  \usepackage{pgfplots}
  \pgfplotsset{compat=newest}
  \usetikzlibrary{plotmarks}
  \usetikzlibrary{arrows.meta}
  \usepgfplotslibrary{patchplots}
  \usepackage{grffile}
  \usepackage{amsmath}
  \usepackage{placeins}
  \usepackage{float}
  
    \pgfplotsset{plot coordinates/math parser=false}
    \newlength\figureheight
    \newlength\figurewidth

\usepackage{algorithm}
\usepackage{algorithmicx}

\usepackage[disable]{todonotes}
\usepackage{lipsum}

\usetikzlibrary{external}
\journal{Robotics and Autonomous Systems}

\begin{document}

\begin{frontmatter}



\title{A Comparative Study of Bug Algorithms for Robot Navigation}


\author{K. N. McGuire$^{1*}$, G.C.H.E. de Croon$^1$ and K. Tuyls$^{2}$}

\address{$^1$Delft University of Technology, The Netherlands\\
	$^2$University of Liverpool, United Kingdom\\
	*Corresponding Author: k.n.mcguire@tudelft.nl}

\begin{abstract}
This paper presents a literature survey and a comparative study of Bug Algorithms, with the goal of investigating their potential for robotic navigation. At first sight, these methods seem to provide an efficient navigation paradigm, ideal for implementations on tiny robots with limited resources.  Closer inspection, however, shows that many of these Bug Algorithms assume perfect global position estimate of the robot which in GPS-denied environments implies considerable expenses of computation and memory -- relying on accurate Simultaneous Localization And Mapping (SLAM) or Visual Odometry (VO) methods. We compare a selection of Bug Algorithms in a simulated robot and environment where they endure different types noise and failure-cases of their on-board sensors. From the simulation results, we conclude that the implemented Bug Algorithms' performances are sensitive to many types of sensor-noise, which was most noticeable for odometry-drift. This raises the question if Bug Algorithms are suitable for real-world, on-board, robotic navigation as is. Variations that use multiple sensors to keep track of their progress towards the goal, were more adept in completing their task in the presence of sensor-failures. This shows that Bug Algorithms must spread their risk, by relying on the readings of multiple sensors, to be suitable for real-world deployment.

\end{abstract}


\end{frontmatter}

%\tableofcontents

\section{Introduction}\label{sec:othernavigationmethods}

Robotic indoor navigation of robots has been a sought-after topic for the last few decades within the robotic community. An important stimulus for this interest is its potential for a wide range of scenarios, e.g. search-and-rescue, greenhouse observation, industrial inspection. Indoor navigation also comes with a wide range of issues, such as the absence of a reliable GPS-signal and wall interference in long-range communication. An indoor robot should preferably be autonomous and able to navigate based on its on-board sensors and computational capacity.

There has been tremendous progress in autonomous robotic navigation, up to a point that some researchers believe this to be an already solved problem. With the emerging autonomous cars, simultaneous localization and mapping (SLAM) has reached high maturity in development (see \cite{bresson2017simultaneous} for a review).  SLAM is a notoriously  complex and expensive algorithm, consuming much of the robot's on-board progressing power. To strive towards computationally efficient methods is advantageous for any robot, but it becomes  vital when the application requires the use of tiny, light-weight robots. For instance, small Micro Aerial Vehicles (MAVs), in the order of 50 grams and 15 cm diameter, could be ideal for exploring small and confined spaces. However, their on-board computational resources are so limited that currently they cannot make use of SLAM methods. 
\begin{figure}[t]
	\footnotesize
	\centering
	 \includesvg[width=0.25\textwidth]{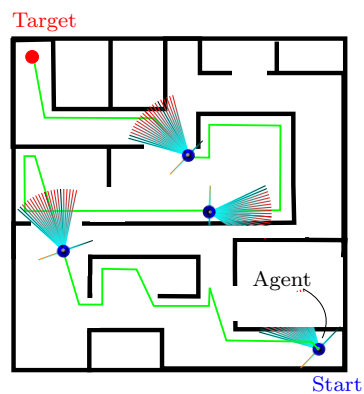}
	\caption{An example of an agent performing a Bug Algorithm-like behavior, while navigation in a indoor environment. From a starting position (bottom-right), it moves towards the target (top-left), where it tries to move towards the target whenever it can, and follows the obstacles' boundary when it hits an obstacle. Its trajectory is given in green.}\label{fig:frontpagebug}
\end{figure}

Given these strict computational requirement for tiny robotic platforms, an important question is raised: does the actual simple principle of navigation, \textit{going from point A to point B}, need the computational and memory requirements for constructing and maintaining high-resolution metric maps? Should the complexity of the strategy not be proportional to the simplicity of the task?

There are several light-weight alternatives to SLAM to consider, such as Topological SLAM (see \cite{Boal2014a} for a review). Biologically inspired techniques like the Snapshot Model (\cite{Cartwright1983}) and the Average Landmark Vector (\cite{lambrinos2000}) can also be considered. These efficient methods, however, still have the tendency to scale up the memory requirements, when navigating in a more complex and large environments.

 In this article, we will look at a navigation method of a different kind: \emph{Bug Algorithms}. Although the name suggests a biological origin, it is a path-planning technique that evolved from maze-solving algorithms. The main principle of Bug Algorithms is that they do not know the obstacles in their environment and only know their target's relative position. They will locally react only upon contact with obstacles and walls, in a way that lets the agents progress towards their goal, by following the obstacle's boundaries ("wall-following"), as illustrated in Fig.~\ref{fig:frontpagebug}. The nature of Bug Algorithms is ideal for indoor navigation on tiny, resource-limited, robotic systems, as their potential memory and processing requirements are low, therefore expected to take up little space on the on-board computer. This will free up resources for other tasks/behaviors.

In this paper we will delve into Bug Algorithms in more detail, by providing an overview of the techniques existing today. There have been two comparative studies on Bug Algorithms before (\cite{Noborio2000}, \cite{Ng2007}), however the biggest difference is that we will evaluate how suitable Bug Algorithms are in becoming a new navigation standard within robotics. Here we will look into the assumptions made about the environment and we will evaluate whether they are realistic. An important conclusion of our study is that Bug Algorithms tend to heavily rely on a perfect position estimation, which cannot be taken for granted in a GPS-deprived indoor environment. Global positioning systems could be set up beforehand, such as a motion capture or Ultra-Wide-Band (UWB) localization system (like in \cite{mueller2015fusing}). However, in cases like search-and-rescue scenarios, it is undesirable to have humans prepare the robot's environment. The robots would need to rely on their estimated position, obtained by their own, noisy, on-board sensors.

We will compare a representative subset of Bug Algorithms in the ARGoS simulator, which is capable of modeling realistic physical interactions with objects in the environment. Although we will not implement as many Bug Algorithms as \cite{Ng2007} did, we will test them in more realistic real-world conditions, containing elements such as odometry-drift or recognition-failures. We investigate their behaviors on hundreds of procedurally generated indoor environments, to compare their performance statistically. Here it is shown that the increased measurement noise on the on-board sensors causes a dramatic drop in overall performances of the Bug Algorithms. We will discuss how this affects the potential of Bug Algorithms in robotic navigation and what type of assumptions we can make about the environment, which can point us to the variations that are the most suitable.

An overview of Bug Algorithms is given in section~\ref{ch:bugalgorithms}, starting from their "maze-solver" origins, to the fundamental contact-based Bug Algorithms, to the more recent extended range-based versions and hybrid solutions. This is followed by a sum-up of the methods already used in robotic navigation in section~\ref{ch:implemetation}. Subsequently, we perform
a quantitative comparison of the Bug Algorithms performances, of which the setup is explained in section~\ref{ch:experimentalsetup}. The experiments themselves are discussed in section~\ref{ch:experimentresults}, and involve various degrees of sensor-noise and -failures. The findings of this paper will be discussed in section~\ref{ch:discussion}, from which we will present our conclusions in section~\ref{sc:conclusion}.

\section{Theory and Variants of Bug Algorithms}\label{ch:bugalgorithms}

The late 80s is when the term \emph{Bug Algorithms} first came into existence, evolving from the existing path planning algorithms like Dijkstra (\cite{dijkstra1959note}) and A* (\cite{hart1968formal}). However, the latter methods need to know their environment in advance, which includes start and goal positions, all obstacles and their position along the way. With this information, they need to find the quickest path from A to B within a predefined scenario\footnote{Also called the "piano movers problem"}. But yet, what if the location, size, shape and the amount of those obstacles are not known?

\subsection{Origins: Maze solving algorithms}

Maze-solving algorithms first explored the navigational problem without knowledge about the environment, for enclosed spaces with walls and only one entrance and exit.  The \textit{random-walker} algorithm is the  simplest technique to solve a maze (\cite{evans2017optimization}). It moves in a straight line until it encounters an obstacle. At that point, it will choose an arbitrary and oblique direction to go to next. Luck determines the random walker's success and it could take a significant amount of time before the exit is reached. 

If the target is reachable through a series of interconnected walls,  a \textit{wall-follower} would guarantee a quicker solution than the random mouse (\cite{mishra2008maze}). Its left or right side must be in contact with the boundary of the obstacle or wall while it moves towards the exit. However, if the environment is not an interconnected maze and contains disjoint obstacles between the start- and end-location, the wall-follower might get stuck in an endless loop.

The \textit{Pledge} algorithm can handle a maze with disjoint walls  (\cite{Abelson1986})\footnote{The Pledge algorithm was originally intended as a mathematical educational tool}. The Pledge-agent will first commit ("pledge") to a fixed oblique direction in heading and moves there in a straight line. If it hits an obstacle, it will adapt a wall-following behavior, while monitoring the changes in heading. If the angular sum of its heading, with respects to its initially committed heading, returns to 0\degree (here not equivalent to 360\degree),  the Pledge algorithm will leave the obstacle at that point and continue to follow the original direction it started out with. This enables the Pledge agent to also handle mazes with disjoint walls, which is an improvement from the simple wall-follower. However, this algorithm will by itself not move directly towards the exit, as it does not have any knowledge of where it is. If, for instance, its final goal is a fixed position located in an wide open space, the Pledge-algorithm could miss it entirely.

\begin{figure}
	\centering
	\footnotesize
	\includesvg[width=0.5\textwidth]{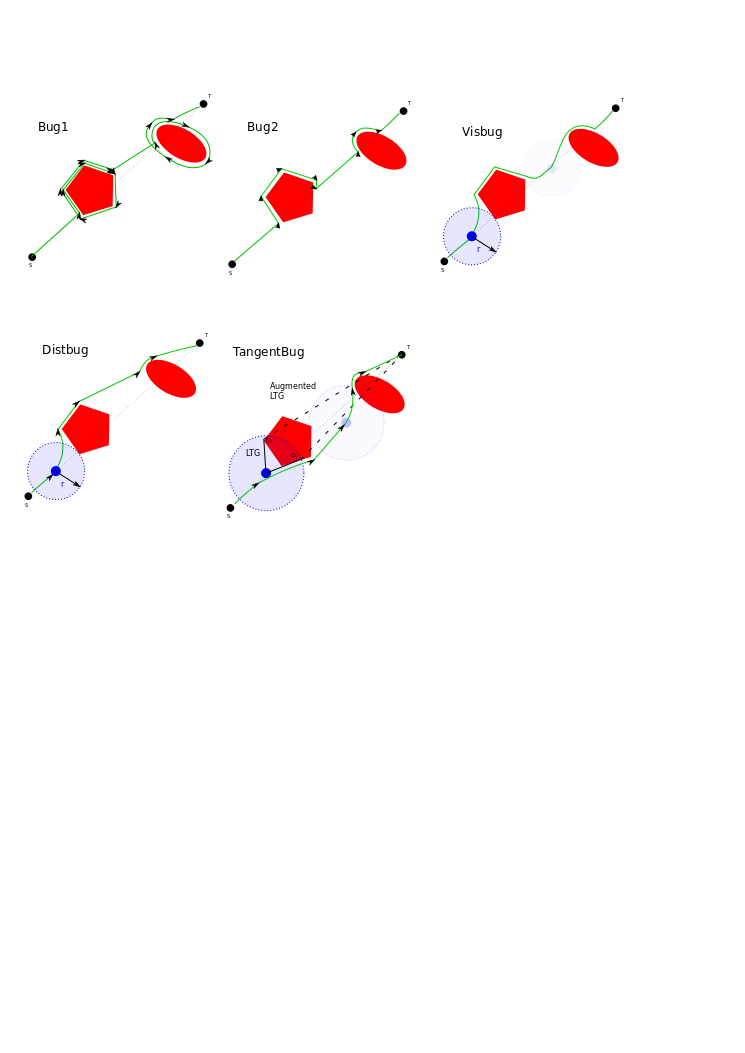}
	\caption{The behavior of simple Bug Algorithms: a) Com, b) Bug1 and c) Bug2. The \textit{S} and \textit{T} depicts the start and target position respectively.}\label{fig:simplebug}
\end{figure}

\begin{figure}
	\tiny
	\centering
	\includesvg[width=0.5\textwidth]{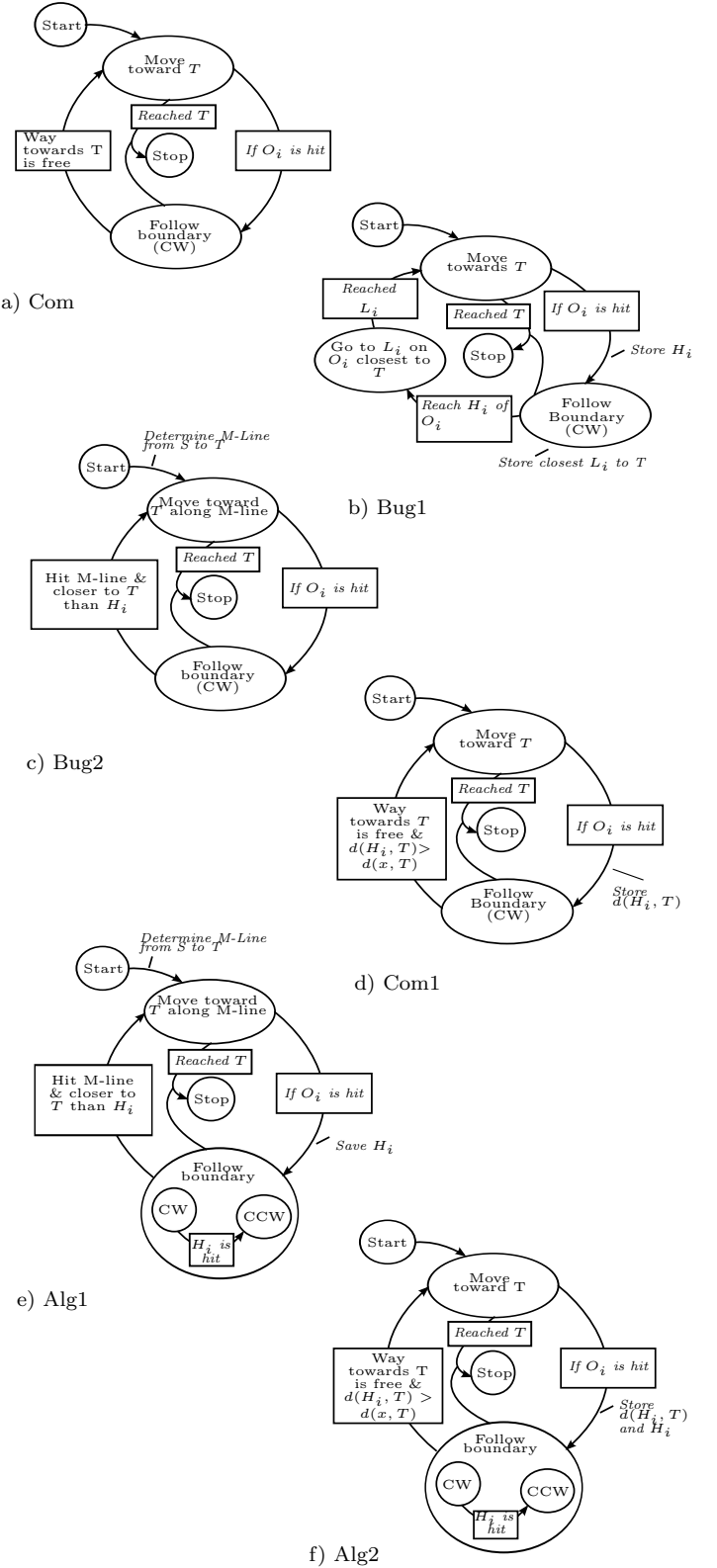}
	\caption{Bug algorithm state machines. The \textit{S} and \textit{T} represent the start and target position respectively. $O_i$ is the i-th obstacle that the bug hits and $L_i$ and $H_i$ is the i-th leave- and hit-point, respectively. 
	}\label{fig:simplebug_statemachine}
\end{figure}

\begin{figure}[t]
	
	\centering
	\tiny
	\includesvg[width=0.5\textwidth]{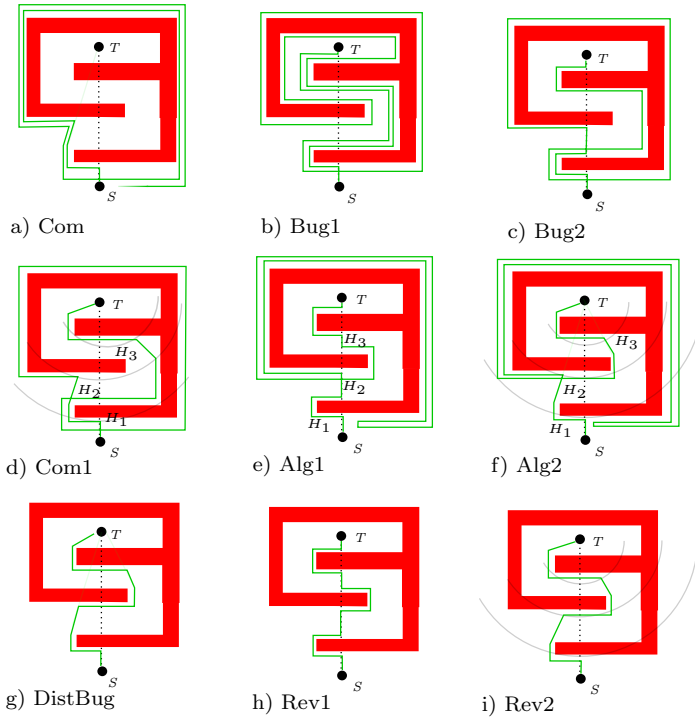}
	\caption{Generated paths by the Bug Algorithms (a) Com, (b) Bug1, (c) Bug2, (d) Com1, (e) Alg1, (f) Alg2, (g) DistBug, (h) Rev1 and (i) Rev2 in a more challenging environment. The \textit{S} and \textit{T} depicts the start and target position respectively and $H_i$ means the i$^{th}$ hit-point. x is the current position of the agent and CW and CCW stand for Clock Wise and Counter Clock Wise respectively.}\label{fig:simplebug_complexenvironment}
\end{figure}

\subsection{Contact  Bug algorithms}\label{sec:contactbug}
%explain hit and leave points
Typical indoor environments have corridors, rooms and disjoint obstacles, where Bug Algorithms (BAs) should be able to solve the path planning problem. \citet{Lumelsky1986} are  the pioneers in developing this new technique. At first, they described a very simplistic BA, called the "common sense algorithm" which can be abbreviated as \textit{Com}.  Fig.~\ref{fig:simplebug_statemachine}(a) shows a state machine of the BA, where it will move towards the target whenever it can. This results in the behavior illustrated in Fig.~\ref{fig:simplebug}(a). The position where a BA hits the obstacle for the first time is called a \emph{hit-point}, and it has a \emph{leave-point} as soon as the direction to the target is free. Intuitively, Com  could solve many situations; however, \cite{Lumelsky1986} pointed out that there are scenarios in which it cannot reach the goal. This happens when introduced to a more complex environment as, for instance, the one illustrated in Fig.~\ref{fig:simplebug_complexenvironment}(a).

In the same paper of \cite{Lumelsky1986}, the \textit{Bug1} algorithm was introduced, following a different strategy to overcome the issues that Com is facing (see Fig.~\ref{fig:simplebug_statemachine}(b) for its state machine).  Every obstacle Bug1 comes across, it first has to "explore" the obstacle by following its entire border, while simultaneously keeping track of which position is the closest to the target, as shown in the simple environment in Fig.~\ref{fig:simplebug}(b). After it encounters its original hit-point, Bug1 will continue and move towards the position closest to the target, from which it will leave the obstacle. The path length will therefore never exceed the limit: $P~=~d(S,T) + 1.5 \cdot \sum p_i$, where P is the total path length, $d(S,T)$ the direct distance between the start (S) and target (T) location and $p_i$ is the length of the boundary of the i$^{ th} $ obstacle. Bug1 is able to handle environments where Com failed (as seen in Fig.~\ref{fig:simplebug_complexenvironment}(b)); however, it is a less intuitive approach. As it needs to know the entire border of the obstacle, this will naturally create unnecessary long paths.

\citet{lumelsky1987path} recognized the non-optimal path-lengths of Bug1, and therefore proposed an alternative:\textit{ Bug2}. Between the beginning and end position, an imaginative line is drawn, called the \emph{M-line} (see Fig.~\ref{fig:simplebug_statemachine}(c) for Bug2's state-machine). In the simple scenario of Fig.~\ref{fig:simplebug}(c), this means that the bug will follow the obstacles border until it hits the same M-line at the other side. As long as that point is closer towards the target than the hit-point's position, it will depart from the obstacle. This reduces the maximum possible BA's path length to  $P~=~d(S,T) + 1 \cdot \sum p_i $, which is also illustrated by Fig.~\ref{fig:simplebug_complexenvironment}(c).

\citet{Sankaranarayanan1990} still  found  scenarios in which Bug2 would still produce an unnecessary long path. According to them, it is because of its incapability to store and compare previous visited points along the obstacle's boundary. They extended the Bug2 algorithm with the following principle: to change its wall-following direction if it comes across a previously visited hit-point along the border of the obstacle. It has been dubbed as \emph{Alg1}, which can be seen in Fig.~\ref{fig:simplebug_statemachine}(d). It is true that in some situations a shorter path will be generated, however Alg1's maximum \textit{possible} path length is longer: $P~=~d(S,T) +  2\cdot \sum p_i$. Fig.~\ref{fig:simplebug_complexenvironment}(e) shows an example of its behavior in a complex environment.

\cite{Sankaranarayanan1990} also expressed interest for the intuitive method Com, as it does not exploit the restrictive M-line, but leaves the boundary as soon as there is a free space between the BA and the target. They suggested an extended version of Com, \emph{Com1}\footnote{This is also being referred to as \emph{Class1} in the studies of \cite{Noborio2000} and \cite{Ng2007}}, which remembers the previous hit-point's distance to the target. Com1 will utilize this as an extra argument in his state-machine (Fig.~\ref{fig:simplebug_statemachine}(d)),  to initiate the departure from the obstacle boundary, as seen in Fig.~\ref{fig:simplebug_complexenvironment}(d).  Based on Com1, \emph{Alg2} was created in the same paper of \cite{Sankaranarayanan1990} as well, where it, similar to Alg1, reverses the wall-following direction if it encounters a previous saved hit-point (Fig.~\ref{fig:simplebug_statemachine}). Alg2 therefore needs to keep track of all previous hit-points on its way for the reverse local direction condition, as it this will occasionally occur (Fig.~\ref{fig:simplebug_complexenvironment}(f)).\footnote{ The statemachine of Com, Com1, Bug2, Alg1 and Alg2 are also available as pseudo code in appendix~\ref{ap:pseudoBAs}, as they will be implemented later in this paper for the comparative study.}

\begin{figure}
	
	\centering
	\footnotesize
	\includesvg[width=0.5\textwidth]{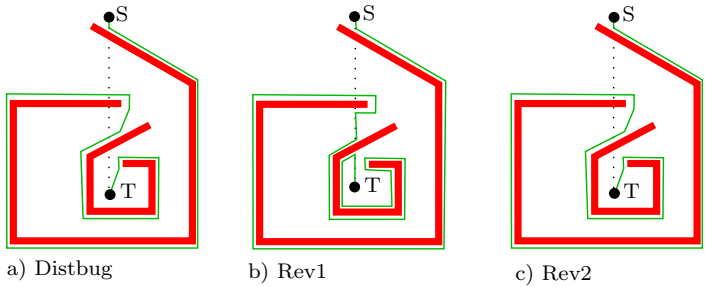}
	\caption{An alternative complex environment to show a case that would produce a long path-length for a) DistBug, b) Rev1 and c) Rev2. }\label{fig:complex_environment_fail}
\end{figure}

\cite{Kamon1997} created a BA quite similar to Alg2, \emph{DistBug}\footnote{Here we are revering to the extended DistBug algorithm of the same paper, with the search manager and local-direction choice based on the slope of the wall.}. The only difference is that it will not remember the positions of all the previous hit-points, but solely the last one, therefore making it more memory efficient. Another intriguing aspect of DistBug, is that there is no fixed initial local wall-following direction along the boundary of the obstacle, as it depends on the orientation on which the BA touches the hit-point. Most times, this will naturally lead it to the target  and result in a shorter path,  which is noticeable in the environment illustrated in Fig.~\ref{fig:simplebug_complexenvironment}(g). However, there are situations where such a policy will fail, as in Fig.~\ref{fig:complex_environment_fail}(a). At the first hit point, it would be better to follow the wall in the other direction.

An extension to both Alg1 and Alg2 was proposed by \cite{horiuchi2001evaluation}, named \emph{Rev1} and \emph{Rev2} respectively. Both BAs will alternate their local direction at each (new) hit-point, which is a good strategy for environments like in Fig.~\ref{fig:simplebug_complexenvironment}(h) and (i). Rev1\&2 save the chosen local direction and its associated hit-point in a list. If these locations are revisited  again, the bug algorithm will chose the opposite local direction than the one stored. However, the "best" choice for the local wall-following direction is not trivial. Fig.~\ref{fig:complex_environment_fail}(b) and (c) show a situation where alternating the local wall-following direction is not the best policy. One may argue that the shown case is disadvantageous to Rev1 and Rev2, as they do not encounter any previous hit-points on their path. However, the examples does show that the best choice of local direction depends on the environment. It is, therefore, difficult to find a generic strategy for determining the best wall-following direction.

\subsection{Bug Algorithms with a Range Sensor} 

\begin{figure}[t]
	\centering
	\footnotesize
		\includesvg[width=0.5\textwidth]{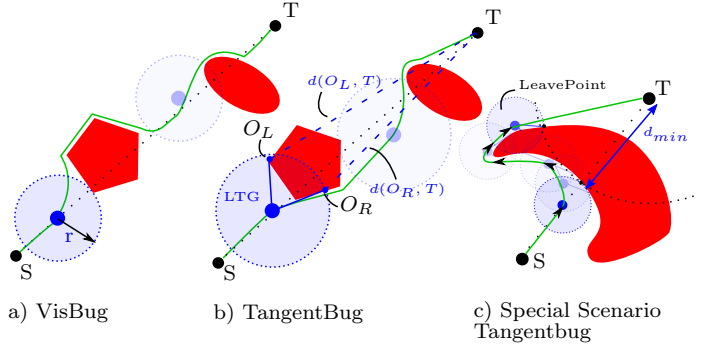}

	\caption{The Bug Algorithms developed with obstacle detection with range sensors: (a) VisBug and (b) TangentBug. The \textit{S} and \textit{T} depicts the start and target position respectively. $r$ stands for radius of the range sensor. LTG stands for local tangent graph and $O_L$ and $O_R$ stand for the left and right border of the detected obstacle within the range sensor respectively. $(d(O_R,T)$ and $(d(O_L,T)$ stand for the distance between the left and right obstacle boundary to the target, respectively. (c) A close up of a scenario in which Tangent bug is able to handle local minima.    }\label{fig:simplebug_range}
\end{figure}

What if the robot is able to sense obstacles already at a certain range and therefore act before touching the obstacles physically? \citet{Lumelsky1986} already mentioned this idea in their first paper, which has been materialized in \cite{Lumelsky1988} and \cite{Lumelsky1990} as \textit{VisBug 21 \& 22}. Both are based on Bug2, but are also equipped with a range sensor able to sense up to a given maximum range. The BA will still follow the M-line but they can detect "short-cuts" to the next obstacle which should reduce the total path traveled,\footnote{No indication of the path length is given here, however, of the complexity, this will no longer be mentioned from here on in this report.} as can be seen in Fig.~\ref{fig:simplebug_range}(a).

\citet{Kamon1996} introduced a successful version of the range-based Bug Algorithms, called \textit{TangentBug}. Within the maximum range of its sensor, a local tangent graph (LTG) is constructed, as illustrated in Fig.~\ref{fig:simplebug_range}(b). The LTG represents the discontinuities/borders of the detectable obstacle field around the robot. It starts out by moving  towards the target while traversing the LTG edge that results in the quickest path to the target ($T$) from its current position ($x$). This goes as follows:
\begin{eqnarray}
i \leftarrow argmin_i (D_i)
 ~\text{with}~\\
 D_i= d(x,O_i)+d(O_i,T) \\
i = left, right
\end{eqnarray}  

, where $D_i$ is the distance  of the agent towards the left or right obstacle $O_i$ ($d(x,O_i)$) plus the remaining distance from that obstacle to the target $d(O_i,T)$. TangentBug will always follow the LTG edge which is expected to result in the smallest path towards the target. However, if $D$ of that edge increases, it will save the current range to the target as a local minimum ($d_{min}$) and will continue following the remaining boundary of that obstacle. If the robot senses a node on the boundary of the obstacle that is smaller than $d_{min}$, it trigger a leave-condition and, if possible, moves directly to the target (see Fig.~\ref{fig:simplebug_range}(c)). \cite{Kamon1999a} extended TangentBug to operate in 3D-environments as well (\emph{3DBug}).

TangentBug is probably the most referred work in the field of BAsand many variants of it exists.  The 360\degree~range sensor assumption is changed to a sensor with a limited field of view with \textit{WedgeBug} (\citet{Laubach1999}), for instance,to represent a stereo camera. \citet{Magid2004} developed a BA which will actively search for the right local wall-following direction, to prevent a long-path length. Their \emph{CautiousBug} will not choose a direction based on the angle of attack on the hit-point, as DistBug, but will first do a spiral search along the border, with the hit-point in the center. A disadvantage of this method seems that the spiral search by itself will also produce a long path, therefore it has less of an advantage over Tangentbug. A newer variation is \textit{InsertBug} by \citet{Xu2013}, which navigates by means of way-points, placed on a safe distance from the obstacle's boundary. This could be seen as a version of  TangentBug that adds a safety margin to each obstacle detected.

\subsection{Special Bug Algorithms}

Some BAs either take a special approach or are combined with other existing methods (\emph{HybridBugs}). \cite{Lee1997} used fuzzy logic with an adjusted Com method, a.k.a \emph{FuzzyBug}. Assuming the BA is equipped with two single-beam sensors, pointed forward on both sides, it can detect if an obstacle is closer to its right or left. Based on a fuzzy membership function, FuzzyBug decides to follow the obstacle's boundary on its right or left, which a similar approach to DistBug's strategy. 

\citet{Noborio1999} developed HB-I, which is another HybridBug. After each hit-point of the obstacle, HB-I moves along the border in both directions until it hits a corner. It will then select the best direction first, based on the best-first search of a decision tree. %The combination with search algorithms makes this method computationally more complex. 
\citet{Xu2014a} used a different approach with \emph{RandomBug}. Once it detects an obstacle, it generates random points within the range of its sensor. From these points, RandomBug selects the optimal one,  dependent on how close the point is to the target, and generates a motion vector towards it. This produces a path quite similar to InsertBug, but the process is highly related to rapidly-exploring random trees (\cite{lavalle2001randomized}).

%As time progresses, the enhancements of Bug Algorithms tend to over-complication. Nevertheless, some still strive to maintain simplicity. 
\citet{Taylor2009} developed \textit{IBug}, which is short for Intensity-bug. Its only information about its target is a wireless beacon on the specified location, of which it will navigate towards by means of the signal strength. Since they assume that IBug can make use of a "tower-orientation" sensor, the agent will move towards the beacon location. When it does, IBug will temporarily save the value of the intensity ($i_H$) at that very moment. Here, a high intensity (signal strength) means a short distance to the target and a low intensity a large distance. While the robot follows the obstacle's boundary (always CW or CCW), it compares the current intensity level with $i_H$, as well as the current intensity and of time-steps back. If the signal strength decreases after increasing, the agent will have detected a local minima and  a leave-condition is triggered, but only if the current measured intensity is larger than $i_H$. Although the leave condition is different, the latter comparison of intensity levels at the hit- and leave-points is quite similar in approach to Com1, with intensities substituted for distances.

\subsection{Overview Bug algorithms}\label{sec:overview}
\begin{figure}[t]
	\footnotesize
	\centering
	\includesvg[width=0.5\textwidth]{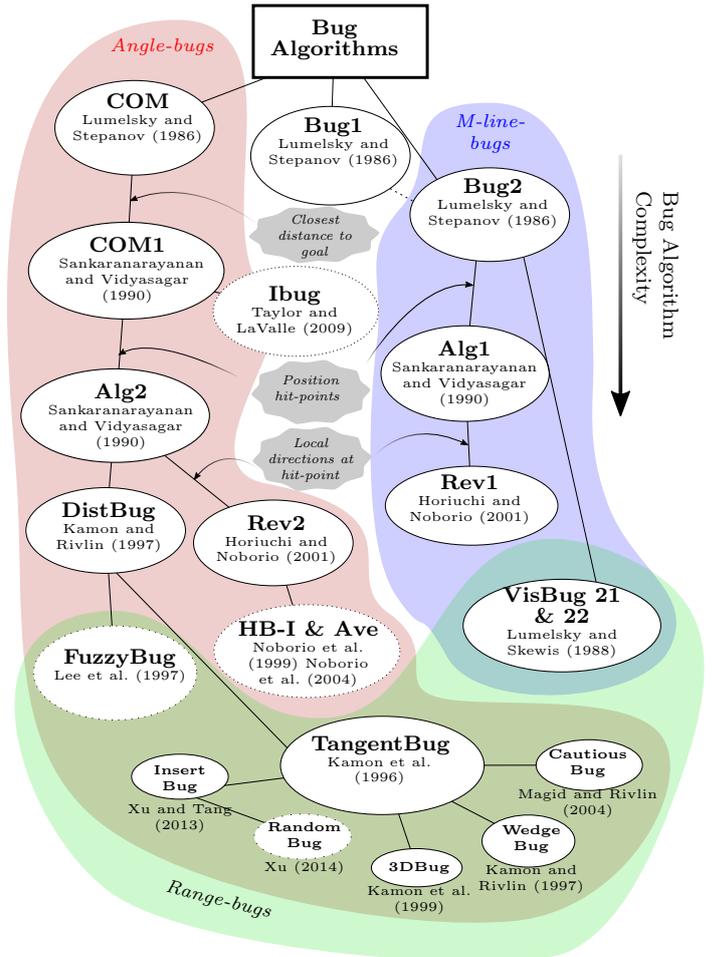}
	\caption{An overview of all the discussed Bug Algorithms (BAs) in section~\ref{ch:bugalgorithms}. These BAs are presented in a development tree of increasing complexity. It makes a distinction between Angle-Bugs, BAs that move to the target's azimuth direction, M-line-Bugs, BAs that use an M-line to navigate, and Range-Bugs, which use a range sensor to detect obstacles. The BAs noted in a dotted circle are special/ hybrid-bugs. The gray blobs indicate the type of memory and leave-condition added to the method. The latter is only shown until Rev1\&2.}\label{fig:scheme_bugs}
\end{figure}

The BAs discussed in the previous sections are visualized in the overview of Fig.~\ref{fig:scheme_bugs}, where they are connected based on their development and features.We subdivide the algorithms in a few major categories. The main division already started in the paper of \cite{Lumelsky1986}, where they presented Com, Bug1 and Bug2. Com led to a series of Bug Algorithms that navigated in an azimuth angle towards the goal  whenever it had the chance to do so.  Hence, here we categorize them 
as \emph{Angle-Bugs}. \cite{Lumelsky1986} realized that their next creation, Bug1, would create long trajectories by default. The community seems to have agreed as no extension or variation of Bug1 was developed here after, so therefore no similar Bug Algorithms has emerged since. \cite{Lumelsky1986}'s alternative solution, Bug2, did have more potential, so new variations of \emph{M-line-Bugs} have been presented, leading to a separate category of BAs.

Com is arguably the most simplest BAs, as it uses no memory, nor determines any M-line. Although for some simple environments this proved to be sufficient, Com has a chance of never reaching its final destination in more complex scenarios. With Com1, \cite{Sankaranarayanan1990} added a distance-based leave condition, where it will only leave the obstacle if it reached a position closer as it has been before. This requires Com1 to remember what its latest closest distance was to the goal and use it in the techniques's leave-condition, which has been adopted by the following BAs ever since. 

\cite{Sankaranarayanan1990}'s Alg1 and 2 are given additional memorization tasks. Not only do they remember the previous minimal distance to the goal, but  all the hit-points' locations in between as well. If Alg1\&2  encounter one of those saved hit-points, they will reverse the local direction of their wall/boundary-following. \cite{horiuchi2001evaluation} went one step further and made the Rev1\&2 remember their last local wall-following direction, together with the local direction chosen at each hit-point, and alternates at each revisit. However, DistBug uses a more memory-friendly approach to determine his local position, which is purely based on the detected slope of the approached obstacle, which sets it apart from Rev2. 

Fig.~\ref{fig:scheme_bugs} shows that BAs started to use range-sensors at one point, creating the \emph{Range-Bugs} category. Visbug21\&22 were able to find shortcuts from the M-line to the obstacle's boundary. Both FuzzyBug and TangentBug used their range-sensor to choose the expected best local direction and guide the obstacle-following behavior and the same holds for the many variants of TangentBug. 

We make some general observations about the overview in Figure 7. Firstly, there are more Angle-bug-based BAs than M-line bugs. This is likely thanks to their more intuitive and less restrictive navigation strategy towards the target. Secondly, more and more features are added to the BAs as time progresses. Each new BA builds on top of an other, adding new leaving conditions and memory capabilities, therefore increasing the bug's complexity in the hope to find more efficient paths. The sole exception is the more recent Ibug,  which is a more recent variation, but is only one step away from Com1 in complexity.

\section{Bug Algorithms for Robotic Navigation}\label{ch:implemetation}
\begin{table*}[t]
	\centering
	\caption{Robotic implementations of various Bug Algorithms (BAs). These are evaluated on the type of platform used, whether the environment was real or simulated and which BA type was used. Moreover, it shows the used local sensors for obstacle detection and the used global sensor for a position estimate.}
	\label{tab:roboticimplemitations}
	\footnotesize 
	\begin{tabular}{|p{0.2\linewidth}|p{0.11\linewidth}|p{0.09\linewidth}|p{0.1\linewidth}|p{0.2\linewidth}|p{0.15\linewidth}|}
		\hline
		\rowcolor[HTML]{C0C0C0} 
		Paper                     & Platform                   & Environment & Bug algorithm               & Local sensors                         & Global sensors                     \\ \hline
		\cite{Kamon1997}          & Wheeled robot              & Real        & DistBug                     & Range sensors                         & Global localization \newline (system not given)    \\ \hline
		\cite{Laubach2000}        & Microrover                 & Real    & RoverBug \newline(wedgebug extended) & Stereo images                         & Guiding operator (First person view)                   \\ \hline
		\cite{Mastrogiovanni2009} & Wheeled robot\newline Hexapod robot & Real        & $\mu$NAV                    & Ultrasound range sensor \newline Wheel odometry             & Azimuth angle by photo diodes\newline(only for hexapod) \\ \hline
		\cite{Zhu2010}            & Wheeled robot              & Real        &  Bug2 and a DistBug variant                      & Laser scanner (180 deg)               & Global localization \newline (system not given)    \\ \hline
		\cite{Kim2013}            & Wheeled robot              & Real        & Tangentbug (adjusted)          & Ultrasound range sensor \newline Wheel odometry & Global localization \newline (system not given)  \\ \hline
		\cite{Taylor2014} &	Wheeled robot&	Real&	Ibug&	Touch sensors&	IR Beacon \\ \hline
		
		\cite{Ebrahimi2014}       & Quadcopter                 & Simulation  & UavisBug                    & Camera                                & Motion Capture System              \\ \hline
		\cite{Gulzar2015}         & Wheeled robot              & Real        & Not Given                          & Ultrasound                            & Motion Capture System              \\ \hline
		\cite{Marino2016}         & Quadcopter                 & Simulation  & Bug2                        & Laser scanner (180 deg)               & UWB localization                   \\ \hline
	\end{tabular}
\end{table*}

The BAs presented in the last section are considered as a potential new robotic path planning paradigm,  because of their simplicity and low memory footprint.  We first will discuss how the principle of BAs translates to realistic operating conditions. Afterwards, existing BA robotic implementations will be presented, discussing how well these studies represent real-world scenarios.

\subsection{Bug Algorithms in Real-World Conditions}\label{sec:BArealworld}

In the earlier literature overview in section~\ref{ch:bugalgorithms}, it seems to be the case that BAs heavily rely on perfect localization. They almost all assume that the BA does not know the exact location and shape of the obstacles, however they almost all need to know the exact coordinates of their goal position and their own position. The latter is used for more aspects of BAs than first meets the eye:
\begin{itemize}
	\item  Angle Bugs (i.e. Com, Com1, Bug2, Alg2) need to know the distance and azimuth angle to the target at any point.
	\item M-line-Bugs (i.e. Bug2, Alg1, Rev1) both remember the exact line (and direction) between the starting position and the goal, and recognize if they have reached it.
	\item Hit-point memorizing BAs (i.e Alg1\&2) need to match their current position estimate with previously 
	hit-point coordinates.
\end{itemize}

In a typical indoor GPS-deprived environment, obtaining and maintaining a world position is a significant challenge.. An external global localization system can be set up (e.g. motion capture, UWB triangulation); however, in many scenarios (e.g., in a search-and-rescue scenario) there might not be the possibility or time to do this. Real-world robots will need to rely on odometry, which is prone to errors and has the tendency to drift in time from the ground truth. With ground-bound robots, wheel slippage (\cite{borenstein1996measurement}) can cause an increasing error between the real and estimated position. The same goes for visual odometry (\cite{scaramuzza2011visual}), used by MAVs or hovercraft-like vehicles, where the error of the noisy velocity estimate will get accumulated over time. This is  especially the case in a texture-poor environment.

Some BAs (i.e. Alg1\&2 and Rev1\&2) have to remember the exact coordinates of where they have been, which ensures a convergence to the target.   From a practical perspective, this means that the robot needs to recognize where it has been before. As stated earlier, this could be done by odometry. On the other hand, a BA can recognize its current position with the features of its surroundings. An omni-directional camera can observe the scene which will be memorized with local feature descriptions as SIFT (\cite{Goedeme2007}) or global scene descriptors as Bag-of-Words (\cite{Fraundorfer2007}). It then has to evaluate that template of features the entire time while it is traversing along the border of an obstacle. As with visual odometry, the descriptor's performance depends on the texture of the environment. Practically, this will take up extra capacity of the on-board computer. On a limited platform,  this is something that  is best avoided. Moreover, this principle tends to move towards the definition of map-based navigation.

Most BAs use a Distance-to-Target (DT)  measurement in their leave-condition. Next to using the drift-susceptible odometry, they could also retrieve the DT in ways such as received signal strength intensity (RSSI) of BlueTooth (\cite{bargh2008indoor}) or Ultra-Wide Band (UWB, \cite{guo2017ultra}). This does of course require the placement of a wireless transmitter at the target location.
 Moreover, none of these sensors are perfect. DT estimation by BlueTooth RSSI could get an error up to 2 meters and can not practically determine a range from 5 meters on\footnote{This is based on a Bled112 Bluegiga Bluetooth module} (\cite{coppola2018board}), which is influenced by the amount of environment clutter. UWB has better statistics, with a standard deviation of 0.2 meters  and a maximum limit up to 100 meters\footnote{This is based on a DecaWave UWB module in ranging mode}, with less interference from walls and obstacles in between. Even if the distance measurements are very good, the higher energy expenditure of the latter could be a valid reason to prefer the more energy efficient BlueTooth dongle.

\subsection{Existing Implemented Bug Algorithms for Robotic Navigation}


This section will look at current robotic BA implementation, either in a real world environment or a simulated scenario. An overview of these methods is presented in Tab.~\ref{tab:roboticimplemitations}, which lists the platform they used and shows the sensors the robot was equipped with for local obstacle sensing and global position estimations.

\citet{Kamon1997} were one of the first to consider more realistic  sensors for the agents in BAs. With DistBug, they showed, as one of the first, a BA implemented on an actual wheeled robot, a Nomad200. In their paper they mention that the robot, while moving to the target, only responds to local measurements by the contact sensors. However, the robot always moves towards the target after boundary following, therefore, it must also know its own and the targets position in global coordinates. Although the paper of \cite{Kamon1997} has not specified this, their BA would need to use a global localization system.

\citet{Laubach2000} extended their earlier developed WedgeBug to \emph{RoverBug}, for implementation on a real-world micro-rover. It used a stereo camera to detect and follow the obstacles. However, the initiative to leave the obstacle to move towards the target is controlled by a tele-operator, which monitors the rover through a first-person-view image feed. \cite{Zhu2010}, \cite{Kim2013} and \cite{Gulzar2015} have implemented a BA on autonomous real-world wheeled robots without a tele-operator. In all cases, they were using single beam range sensors and/or a laser scanner. Again, the exact location of the robots is needed in order for the BA to move towards the target. Unfortunately, the papers do not specify which type of global localization system was used in their experiments.

\citet{Mastrogiovanni2009} acknowledged that a robot would not be able to know its exact position, but would need to infer it from its noisy on-board sensors. They  implemented \emph{$\mu$Nav} on a real-world wheeled robot, AmigoBot and a hexapod, Sistino. The first platform used ultra-sonic sensors for obstacle detection and wheel-odometry for global localization. Since the wheeled robot combined its wheel-odometry with the azimuth angle toward the target, it could reach the target location from one room to another, even if the orientation was manually perturbed by the researchers. However, the operation area only spanned across a few rooms and no notion was given of what the navigational limit was, based on accumulated errors of odometry drift. Their second robot, the hexapod, was not able to use odometry, so the azimuth angle had to be given by an external source through photo diodes. 

\cite{Taylor2014} implemented IBug on a small wheeled robot for several small-scale environments. In their previous work (\cite{Taylor2009}), they described the BA to be suitable for navigating towards a single wireless beacon. Nevertheless, for the test on a real robot, a Lego-Mindstorm-based platform, \cite{Taylor2014} used an infra-red (IR) beacon instead. It proved to be challenging for their tests to use the signal strength of i.e. a WiFi beacon at a large range.  The use of the IR beacon did necessitate a constant line of sight, which required the obstacles and walls to be lower than the robot itself. Moreover, the IR sensor could detect a low-resolution bearing towards the beacon, but not the distance towards it. This means is that the minimal-distance-based leave-condition from IBug could not be used. Although the tested environments did not  require this extra argument, it will be essential once loop-detection is required in more complex environments.

 \citet{Marino2016}, from the same group as \cite{Mastrogiovanni2009}, created a simulation of a MAV to navigate through multiple floors. Bug2, enhanced with a  potential-field-based boundary following, is implemented on the simulated quadcopter. The model was equipped with a 360$^\circ$ laser scanner and a salient cue sensor, which is used to detect the target. For simulation it was assumed that its exact location is known, referring to recent UWB localization systems. Moreover, if the agent believes it is at the right goal position but on a different floor, it will use the Dijkstra method to compute the shortest path. This is an interesting choice, as the original Dijkstra algorithm does need to know the grid map of the environment and its obstacles, which is opposite to the problem that BAs intend to solve. 

Another simulated MAV implementation by \citet{Ebrahimi2014} assumes exact localization,  mention a motion capture system. They developed \textit{UavisBug} for a simulated MAV for visual guided navigation. The navigation strategy exists in a 2D horizontal plane only and is quite similar to Bug2. However, they combined the BA with SLAM for the obstacle detection and boundary following, from which they used a potential force field to navigate around the obstacle. Even though, \cite{Ebrahimi2014} and \cite{Marino2016} acknowledged the limited sensing, computing and energy capability of MAVs,  they still combine the efficient BAs with computationally-heavy navigation techniques.

If we look at the existing implementations of BAs in real or simulated robots, they all assume or need explicit global localization,  either by a UWB localization system, a motion capture system or a guiding navigator, for the exception of IBug, which used a visual beacon. \cite{Mastrogiovanni2009} is actually the only one that used the odometry of a (bigger) wheeled robot to recover its own position and to update the azimuth angle towards the target; however, the real-life test was too small to draw any conclusions about the suitability of BAs for robotic navigation. In the comparative study, presented in the next sections, we will test  various BAs with varying amounts of odometry drift, recognition failures and distance noise. This will show that these real-world conditions will have significant effect on the BAs' performances.

\section{Experimental Set-up Comparative Study Bug Algorithms}\label{ch:experimentalsetup}

In this paper, we study whether BAs could be used for real-world robotic navigation. Most indoor environments have more complex obstacle configurations than an open environment with a few convex obstacles. There are many situations where the robots could get stuck on their way, particularly in rooms. In this section, we will present our motivation for this study and the chosen set of BAs to be evaluated. We will then provide the details of the simulation used and the procedural environment generator for typical indoor environments. Afterwards, the implementation specifics of the BAs will be presented, by explaining a wall-following paradigm, which is the foundation for all BAs to be implemented.

\subsection{Motivation and Choice Bug Algorithms}\label{sec:choice}

There have been previous comparisons  between the different BAs. In the paper of \cite{Noborio2000}, Class1, Bug2, Alg1, Alg2 and HB-I, of which the latter is of their own making, were compared and evaluated on their generated path-length within a complex maze. Evaluating four different starting positions, they concluded that Class1 and Bug2 had the longest path-length and usually could not complete the task within the required amount of time. Alg1, Alg2 and in particular HD-I, had shorter path lengths and all finished in time. However, they only based their observations on just one indoor environment.

A newer comparative study was performed by \cite{Ng2007}, on: Bug1, Bug2, Alg1, Alg2, DistBug, TangentBug, OneBug, LeaveBug, Rev1, Rev2 and Class1. They presented the BAs with four types of environments and recorded the total path length for each run. They concluded that in 3 of the 4 environments, Bug1 is the one with the longest trajectory and Tangentbug is the fastest out of the 4. However, for the other BAs, their performance could not be adequately compared due to the inconsistent results.

Here, we test the BAs in hundreds of procedurally generated environments, so we can statistically evaluate their performances. Moreover, we also want to select a set of BAs, from the literature review in Sect.~\ref{ch:bugalgorithms}, to be implemented in a more realistic simulation environment. The selection will not be as large as the selection of \cite{Ng2007} and \cite{Noborio2000}, as we believe that these will have similar effects on BAs that stem from the same groups in the overview shown in Fig.~\ref{fig:scheme_bugs}. In the overview of the BA-methods existing today (section~\ref{sec:overview}), it can be seen that many of the methods are natural increments of one another with increasing complexity.  If the fundamental BAs can be tested with these real-world conditions, we would automatically find the issues that their descendants are facing as well.
 
 \begin{figure}[t]
 	\centering
 	\footnotesize
 	\includesvg[width=0.5\textwidth]{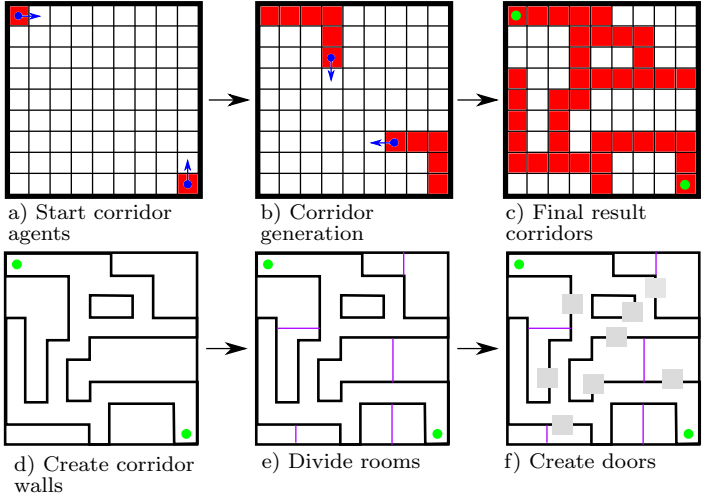}
 	\caption{The steps of the procedural generated environment method will be explained here.  The corridor-generating random agents (blue circles) start in (a) at the same positions as the start and target locations of the experiment. These will move forward in (b), while occasionally turning left and right, while leaving a corridor trace (red blocks). Once it reaches the corridor-density threshold in (c), the corridors-cells are tested for interconnectivity, such that the target position can be reached from the starting position (green circles). The corridor walls are created in (d) and then, in (e), remaining non-corridor spaces are then divided into rooms (purple stripes) and  random  door-openings (gray blocks) are created along the border of the  corridors in (f).}
 	\label{fig:randomenvironment}
 \end{figure}
 \begin{figure}[t]
 	\centering
 	\subfloat[Generated environment in ArGos]{\includegraphics[width = 0.7\linewidth]{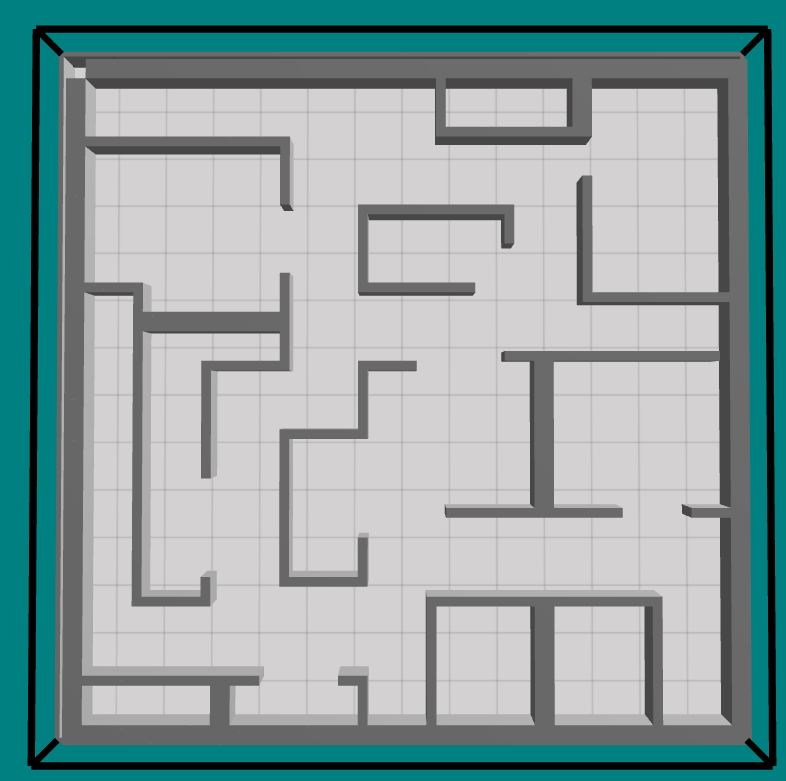}}
 	
 	\vspace{5mm}
 	
 	\subfloat[Modified ArGos Foot-bot]{\includegraphics[width = 0.7\linewidth]{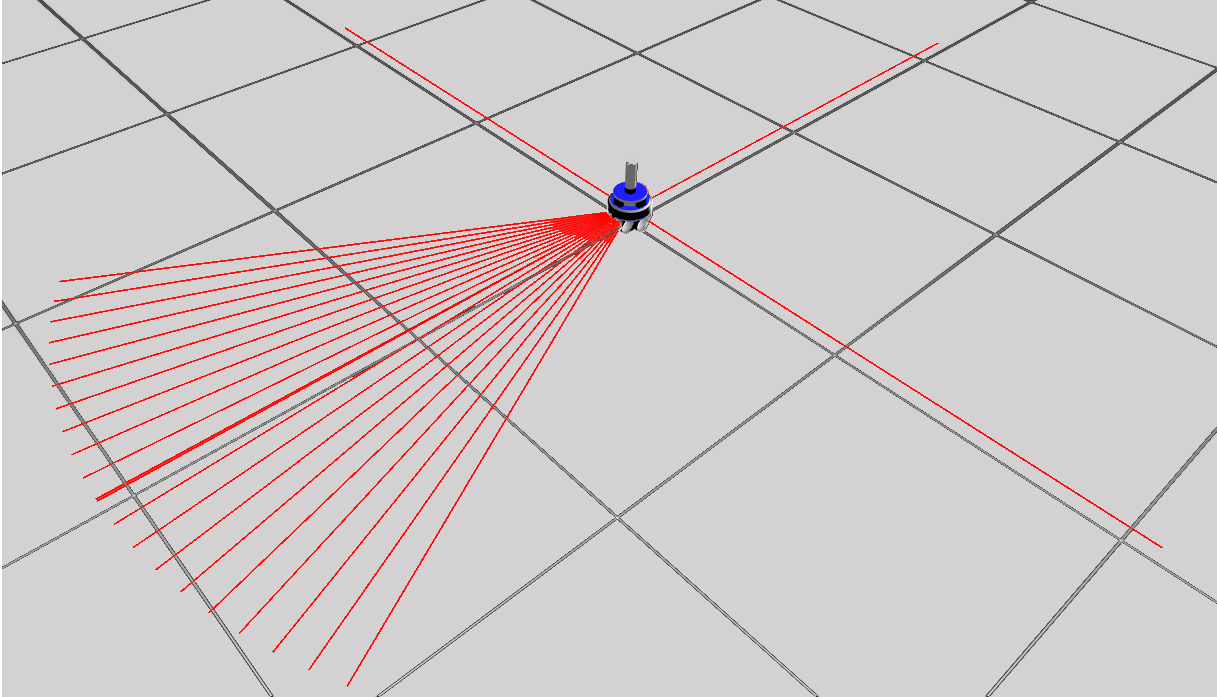}}
 	
 	\caption{(a) Th resulting environment from Fig.~\ref{fig:randomenvironment}(f) generated within the ARGoS simulator  and (b) a modified Foot-bot simulated robot with range sensors (red lines) used for wall following.}\label{fig:randomenvironmentargos}
 \end{figure}
Specifically, we have selected Com, Com1, Bug2 and Alg1\&2, based on the overview in Fig.~\ref{fig:scheme_bugs}. Range-bugs will not be considered as these BAs are the base of those more complex versions. Moreover, the selected BAs presents a mix of different types of strategies (Angle-Bugs and M-line-Bugs) and memory-use (distance and/or hit-points). We will exclude bugs that determine a local wall-following direction, as the policy for this choice is heavily influenced by the structure of the environment, as previously discussed in section~\ref{sec:contactbug}. Moreover, any special bugs will not be considered as well, since they contain aspects and enhancements that no other BA-related research followed up on.

\subsection{Simulation and Procedurally Environment Generator}

It is our ambition to test the earlier mentioned BAs in a simulator with realistic and swift physics calculations. ARGoS, a multi-physics robot simulator developed by \cite{Pinciroli2012},  is used for our comparative study. Its main trait is its efficiency, which enables the simulator to run many times faster than real-time, which will be essential if the BAs are evaluated in many environments. Although ARGoS does have the capability to incorporate its own, C++ based, controller for the robots, the ROS framework is used to enable Python-based controllers. The ROS messaging system is also ideal to modulate whether a new environment needs to be generated, to vary the measurement noise and select the right bug algorithm. 

\begin{figure*} [t]
	\centering
	\scriptsize
	\includesvg[width=\textwidth]{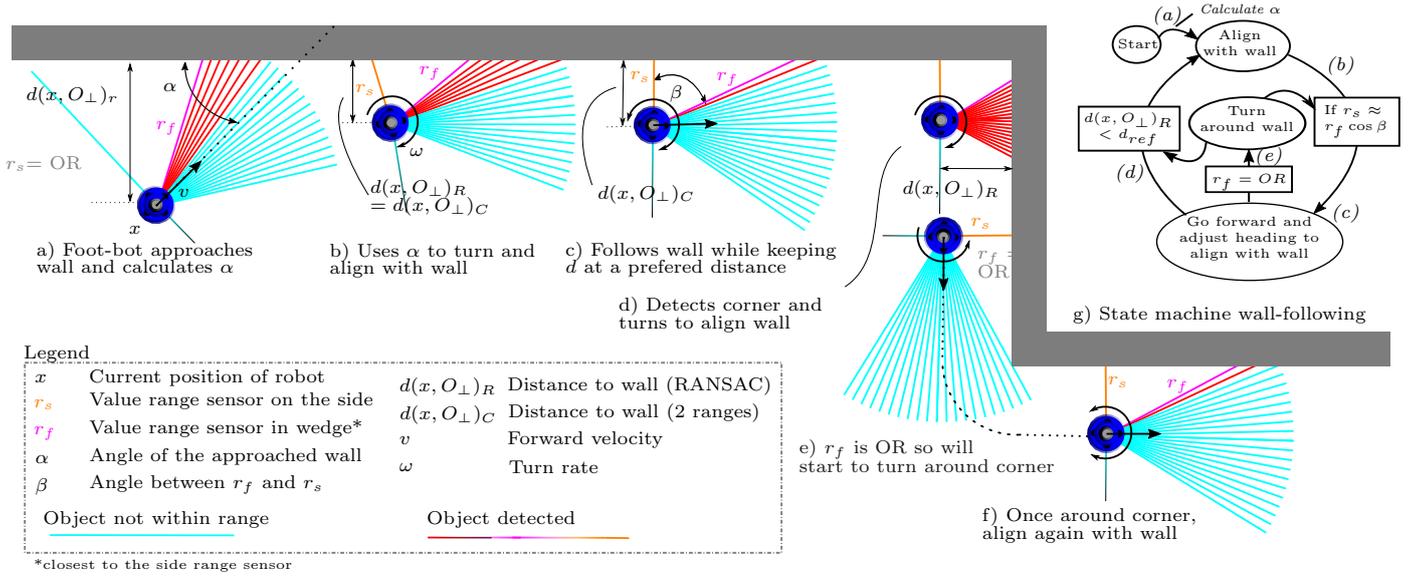}
	\flushleft
	
	\caption{(a-f)Schematics to explain the wall-following paradigm for a right-sided local direction with (g) the  corresponding state machine . OR stand for out of range.}
	
	\label{fig:wallfollowing}
\end{figure*}
\begin{figure} [t]
	%\centering
	%\includegraphics[width = \linewidth]{images/implemented_bug_algorithms.png}
	\footnotesize
	\setlength\figureheight{9cm}
	\setlength\figurewidth{\linewidth}
	\input{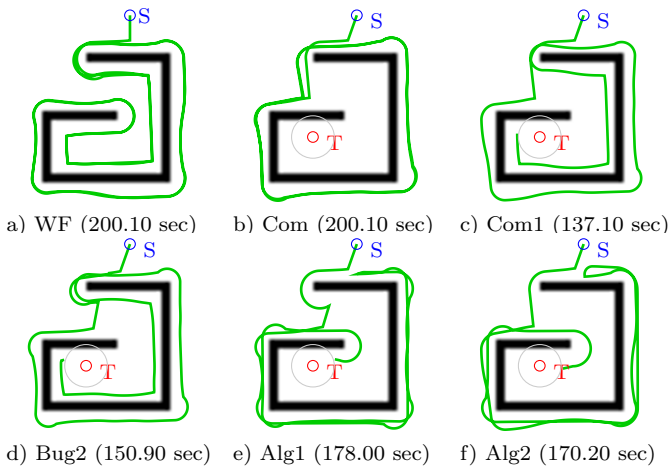}
	
	\caption{The results of (a) the wall-following only (WF) and the implemented bug algorithms that use the same WF (b-e) as part of their navigation strategy. The time limit is 200 sec.
	}
	\label{fig:implementedbugalgorithms}
	
\end{figure}

Since the BAs will be evaluated in many indoor environments, it would be unfeasible to design these by hand. Therefore, a procedural indoor environment generator will automatically generate a new arena for the bugs to navigate in. This process is depicted in Fig.~\ref{fig:randomenvironment}. First, in a coarse grid world (a), two entities are initialized on the exact position of the start and target position to be in the eventual task. They will perform a simple 4-connected path generation, where they will have a certain chance of going straight ($p_{str}$). The chance of either going left or right is equal to $1-p_{str}$. Each agent will leave a corridor trace, as can be seen in (b), until, in (c), the amount of corridors hit a density threshold ($t_{cor}$ = 0.4), which is the  number of grid-cells occupied with a corridor divided by the total number of existing grid cells in the environment. 

A connectivity check is performed, to check if the initial position of the robots are connected by these corridors, which will re-initiate the process in case it fails. This is to ensure that the BA is always able to reach its final destination. Next, walls will be added to these corridors (d). The remaining areas will act as rooms and are divided when they are too large in (e). Finally, in (f), random openings are added along the border of the corridors to create passages these areas. Rooms are especially challenging, as they can lead to agents getting stuck in loops, which will showcase the strengths and weaknesses of the evaluated BAs. The resulting environment in the ARGoS simulation is shown in Fig.~\ref{fig:randomenvironmentargos}(a).

The ARGoS FootBot is used for our experiments, which is a simulated wheeled mobile platform (see \cite{Pinciroli2012} for specifications). The FootBot contains many options to attach various types of sensors, however for our experiments we will only use the proximity sensors. We adapted FootBot to turn the proximity sensors into single beam range sensors with a maximum measurable distance of two meters, placed in the configuration shown in Fig.~\ref{fig:randomenvironmentargos}(b). The robot has two separate single beam range sensors located on each side and 20 range sensors pointing to the front in a wedge shape. This is to simulate a depth sensor/stereo-camera for obstacle detection with a few additional range sensors on the side. Since the robot must move towards the range-wedge configuration, its movement will be non-holonomic.

\subsection{Implementation Details Bug Algorithms}

The most important element of any BA, is the ability to follow a boundary of an obstacle or wall. Based on the  robot configuration in Fig.~\ref{fig:randomenvironmentargos}(b), we developed a wall following principle as illustrated in Fig.~\ref{fig:wallfollowing}. Fig.~\ref{fig:wallfollowing}(a-f) shows the wall following of the footbot for a right-sided local direction and Fig.~\ref{fig:wallfollowing}(g) shows the state machine, which can also be found as pseudo code in appendix \ref{ap:wallfollowing}, Alg.~\ref{alg:wallfollowing}. If the robot moves forward and hits a wall, like in Fig.~\ref{fig:wallfollowing}a, the angle of the wall can be estimated by using a RANSAC line-fit method (\cite{fischler1987random})  in the wedge of range sensors.\footnote{Since RANSAC uses random samples to determine the slope of the plane, some stochastic is expected in the wall-following behavior.} This is done so that the true distance $d(x,O_\perp)_R$ can be estimated from the robot to the wall.\footnote{This only goes with the assumption that the robot will always encounter walls and no single  objects, such plants. The later will not be simulated in ARGoS; however, a classifier able to distinguish walls from small obstacles, must be added if this principle is implemented on a real robot.} If this distance becomes smaller than $d_{ref}$, the preferred distance from the wall, it will keep turning either CW or CCW until it is aligned with the wall. Fig.~\ref{fig:wallfollowing}(b) and (c) shows this alignment for a right-side local direction. This will be the case if the measurement of the side range sensor $r_s$ is equal $r_f \cdot \cos \beta$, where $r_f$ is the element from the range wedge that is the closest to $r_s$ and $\beta$ is the angle between $r_s$ and $r_f$.

After the robot is aligned, it will need to follow the wall, as in Fig.~\ref{fig:wallfollowing}(c). Now the true distance to the wall ($d(x,O_\perp)_C$)\footnote{The $R$ and $C$ subscript of $d(x,O_\perp)$ enables separation of the two methods (RANSAC or only 2 ranges) of retrieving the true distance to the walls.} will be calculated as follows:
\begin{equation}
d(x,O_\perp)_C=\dfrac{r_s\cdot r_f\sin\beta}{\sqrt{r_s^2+r_f^2 - 2\cdot r_s \cdot r_f \cos \beta}}
\end{equation}

The derivation of the latter equation can be found in appendix \ref{ap:realdistance}.  The FootBot will maintain $d(x,O_\perp)_C$ to be near $d_{ref}$, and to keep being aligned in the process. However, since the robot's heading and the measurements of $r_s$ and $r_f$ are coupled, therefore a separate control heuristic is developed to make the wall alignment possible. The details of this wall-alignment method can be found in appendix \ref{ap:wallfollowing}, Alg.~\ref{alg:wallfollowingaligning}.

When the FootBot hits another wall during its forward motion, as in the corner in Fig.~\ref{fig:wallfollowing}(d), while in its wall-following state, it will turn away from the wall until it is aligned with the wall (similar condition as with Fig.~\ref{fig:wallfollowing}(b)). If during a forward motion, the front-range sensor is out of range, as in Fig.~\ref{fig:wallfollowing}(e), the foot-bot will  initiate a wide turn, to find the wall on the other side as in Fig.~\ref{fig:wallfollowing}(f). The state macine for the wall-following behavior can be found in Fig~\ref{fig:wallfollowing} (b),  of which the pseudo code can be found in appendix~\ref{ap:wallfollowing}, Alg.~\ref{alg:wallfollowing}.

This control heuristic  should result in a robust wall-following behavior, in particular for indoor environments with straight walls. The resulting wall-following behavior is shown in Fig.~\ref{fig:implementedbugalgorithms}(a). Here it can be seen that the wall-following produces a smooth path all along the walls of the mirrowed "G". All the implemented bug algorithms, from which the pseudo-code can be found in appendix~\ref{ap:pseudoBAs}, will make use of this exact same wall following behavior in their state machine.  The resulting trajectories in the ARGos simulated environment are shown in Fig.~\ref{fig:implementedbugalgorithms}(b-f).

\section{Experimental Results of Bug Algorithms in Real-World Conditions}\label{ch:experimentresults}

In this section, the BAs will be compared against each other on a wider range and variety of environments than in previous studies. Moreover, we will investigate how sensitive these algorithms are to real-world conditions,  subjecting them to the experimental setup explained in section~\ref{ch:experimentalsetup}. First the selected BAs, which are Com, Com1, Bug2, Alg1 and Alg2 (see subsection~\ref{sec:choice} for the choice's motivation), will be evaluated with perfect localization. After that, the BAs will be subjected to  increasing severity of odometry drift. Next, we will experiment with varying hit-point recognition failures
 and Distance-to-Target (DT) noise. The results of this section will be discussed in the following part of this paper.

\subsection{Experiments with Perfect Localization}

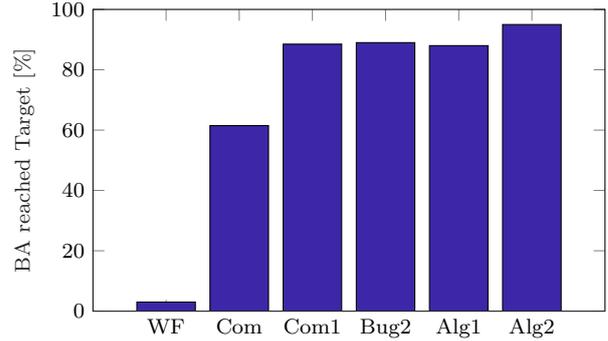
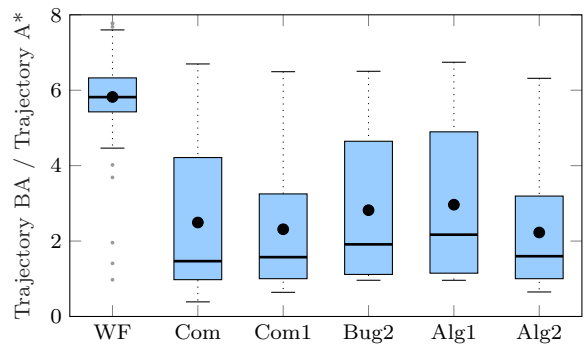
\begin{figure} [t]

	\footnotesize
		\subfloat[Success percentage.]{\setlength\figureheight{4cm}
			\setlength\figurewidth{0.8\linewidth}
\definecolor{mycolor1}{rgb}{0.24220,0.15040,0.66030}%
\begin{tikzpicture}

\begin{axis}[%
width=0.951\figurewidth,
height=\figureheight,
at={(0\figurewidth,0\figureheight)},
scale only axis,
bar shift auto,
xmin=0,
xmax=7,
xtick={1,2,3,4,5,6},
xticklabels={{WF},{Com},{Com1},{Bug2},{Alg1},{Alg2}},
ymin=0,
ymax=100,
ylabel style={font=\color{white!15!black}},
ylabel={BA reached Target [\%]},
axis background/.style={fill=white}
]
\addplot[ybar, bar width=0.8, fill=mycolor1, draw=black, area legend] table[row sep=crcr] {%
1	3\\
2	61.5\\
3	88.5\\
4	89\\
5	88\\
6	95\\
};
\addplot[forget plot, color=white!15!black] table[row sep=crcr] {%
0	0\\
7	0\\
};
\end{axis}
\end{tikzpicture}%
}
		
	\subfloat[Normalized trajectory length.]{
	\setlength\figureheight{4cm}
	\setlength\figurewidth{0.8\linewidth}
\definecolor{mycolor1}{rgb}{0.00000,0.44700,0.74100}%
\definecolor{mycolor2}{rgb}{0.85000,0.32500,0.09800}%
\definecolor{mycolor3}{rgb}{0.60000,0.80000,1.00000}%
\definecolor{mycolor4}{rgb}{0.92900,0.69400,0.12500}%
\definecolor{mycolor5}{rgb}{0.49400,0.18400,0.55600}%
\definecolor{mycolor6}{rgb}{0.46600,0.67400,0.18800}%
\definecolor{mycolor7}{rgb}{0.30100,0.74500,0.93300}%
\definecolor{mycolor8}{rgb}{0.63500,0.07800,0.18400}%
\begin{tikzpicture}

\begin{axis}[%
width=0.951\figurewidth,
height=\figureheight,
at={(0\figurewidth,0\figureheight)},
scale only axis,
xmin=0.5,
xmax=6.5,
xtick={1,2,3,4,5,6},
xticklabels={{WF},{Com},{Com1},{Bug2},{Alg1},{Alg2}},
ymin=0,
ymax=8,
ylabel style={font=\color{white!15!black}},
ylabel={Trajectory BA / Trajectory A*},
axis background/.style={fill=white}
]
\addplot [color=mycolor1, draw=none, mark size=0.5pt, mark=*, mark options={solid, fill=white!60!black, white!60!black}, forget plot]
  table[row sep=crcr]{%
1	7.68040177887169\\
1	7.77242890901024\\
1	7.77555486822652\\
};
\addplot [color=mycolor2, draw=none, mark size=0.5pt, mark=*, mark options={solid, fill=white!60!black, white!60!black}, forget plot]
  table[row sep=crcr]{%
1	0.973583008242971\\
1	1.40835929874253\\
1	1.95658474705115\\
1	3.68665275457326\\
1	4.01854202147537\\
};
\addplot [color=black, dotted, forget plot]
  table[row sep=crcr]{%
1	7.59945295328558\\
1	4.4641950020395\\
};
\addplot [color=black, forget plot]
  table[row sep=crcr]{%
0.86	7.59945295328558\\
1.14	7.59945295328558\\
};
\addplot [color=black, forget plot]
  table[row sep=crcr]{%
0.86	4.4641950020395\\
1.14	4.4641950020395\\
};
\draw[fill=mycolor3, draw=black] (axis cs:0.72,5.42604191102385) rectangle (axis cs:1.28,6.32705708150569);
\addplot [color=black, line width=1.0pt, forget plot]
  table[row sep=crcr]{%
0.72	5.8168532405055\\
1.28	5.8168532405055\\
};
\addplot [color=mycolor4, draw=none, mark=*, mark options={solid, fill=mycolor3, black}, forget plot]
  table[row sep=crcr]{%
1	5.82159334257595\\
};
\addplot [color=black, dotted, forget plot]
  table[row sep=crcr]{%
2	6.69688251866982\\
2	0.387454715460121\\
};
\addplot [color=black, forget plot]
  table[row sep=crcr]{%
1.86	6.69688251866982\\
2.14	6.69688251866982\\
};
\addplot [color=black, forget plot]
  table[row sep=crcr]{%
1.86	0.387454715460121\\
2.14	0.387454715460121\\
};
\draw[fill=mycolor3, draw=black] (axis cs:1.72,0.975885925691737) rectangle (axis cs:2.28,4.21453239524855);
\addplot [color=black, line width=1.0pt, forget plot]
  table[row sep=crcr]{%
1.72	1.46696867909939\\
2.28	1.46696867909939\\
};
\addplot [color=mycolor5, draw=none, mark=*, mark options={solid, fill=mycolor3, black}, forget plot]
  table[row sep=crcr]{%
2	2.49220712628602\\
};
\addplot [color=black, dotted, forget plot]
  table[row sep=crcr]{%
3	6.49136267252451\\
3	0.640645652958544\\
};
\addplot [color=black, forget plot]
  table[row sep=crcr]{%
2.86	6.49136267252451\\
3.14	6.49136267252451\\
};
\addplot [color=black, forget plot]
  table[row sep=crcr]{%
2.86	0.640645652958544\\
3.14	0.640645652958544\\
};
\draw[fill=mycolor3, draw=black] (axis cs:2.72,1.00213549907757) rectangle (axis cs:3.28,3.25034317778108);
\addplot [color=black, line width=1.0pt, forget plot]
  table[row sep=crcr]{%
2.72	1.57303469108231\\
3.28	1.57303469108231\\
};
\addplot [color=mycolor6, draw=none, mark=*, mark options={solid, fill=mycolor3, black}, forget plot]
  table[row sep=crcr]{%
3	2.31313942240909\\
};
\addplot [color=black, dotted, forget plot]
  table[row sep=crcr]{%
4	6.50200422483603\\
4	0.960551756144702\\
};
\addplot [color=black, forget plot]
  table[row sep=crcr]{%
3.86	6.50200422483603\\
4.14	6.50200422483603\\
};
\addplot [color=black, forget plot]
  table[row sep=crcr]{%
3.86	0.960551756144702\\
4.14	0.960551756144702\\
};
\draw[fill=mycolor3, draw=black] (axis cs:3.72,1.1153706926342) rectangle (axis cs:4.28,4.64639332560653);
\addplot [color=black, line width=1.0pt, forget plot]
  table[row sep=crcr]{%
3.72	1.91238121907538\\
4.28	1.91238121907538\\
};
\addplot [color=mycolor7, draw=none, mark=*, mark options={solid, fill=mycolor3, black}, forget plot]
  table[row sep=crcr]{%
4	2.81846105166306\\
};
\addplot [color=black, dotted, forget plot]
  table[row sep=crcr]{%
5	6.74202291205768\\
5	0.959589544688677\\
};
\addplot [color=black, forget plot]
  table[row sep=crcr]{%
4.86	6.74202291205768\\
5.14	6.74202291205768\\
};
\addplot [color=black, forget plot]
  table[row sep=crcr]{%
4.86	0.959589544688677\\
5.14	0.959589544688677\\
};
\draw[fill=mycolor3, draw=black] (axis cs:4.72,1.14806369866847) rectangle (axis cs:5.28,4.89637356710016);
\addplot [color=black, line width=1.0pt, forget plot]
  table[row sep=crcr]{%
4.72	2.16937802023844\\
5.28	2.16937802023844\\
};
\addplot [color=mycolor8, draw=none, mark=*, mark options={solid, fill=mycolor3, black}, forget plot]
  table[row sep=crcr]{%
5	2.96406401139901\\
};
\addplot [color=black, dotted, forget plot]
  table[row sep=crcr]{%
6	6.31578152663007\\
6	0.649797737172833\\
};
\addplot [color=black, forget plot]
  table[row sep=crcr]{%
5.86	6.31578152663007\\
6.14	6.31578152663007\\
};
\addplot [color=black, forget plot]
  table[row sep=crcr]{%
5.86	0.649797737172833\\
6.14	0.649797737172833\\
};
\draw[fill=mycolor3, draw=black] (axis cs:5.72,1.00193948262681) rectangle (axis cs:6.28,3.19321320730075);
\addplot [color=black, line width=1.0pt, forget plot]
  table[row sep=crcr]{%
5.72	1.5982236197501\\
6.28	1.5982236197501\\
};
\addplot [color=mycolor1, draw=none, mark=*, mark options={solid, fill=mycolor3, black}, forget plot]
  table[row sep=crcr]{%
6	2.22710144234832\\
};
\end{axis}
\end{tikzpicture}%

	}

	\caption{(a) The percentage of the wall-follower (WF), and the Bug Algorithms (BAs) Com, Com1, Bug2, Alg1 and Alg2, which made it to the goal in an ideal situation with perfect localization, and (b) the trajectory length normalized by the optimal trajectory length calculated by A*.}

	\label{fig:resultsbugnonoise}
\end{figure}

\begin{figure} [t]

\footnotesize
\setlength\figureheight{9cm}
\setlength\figurewidth{\linewidth}
\input{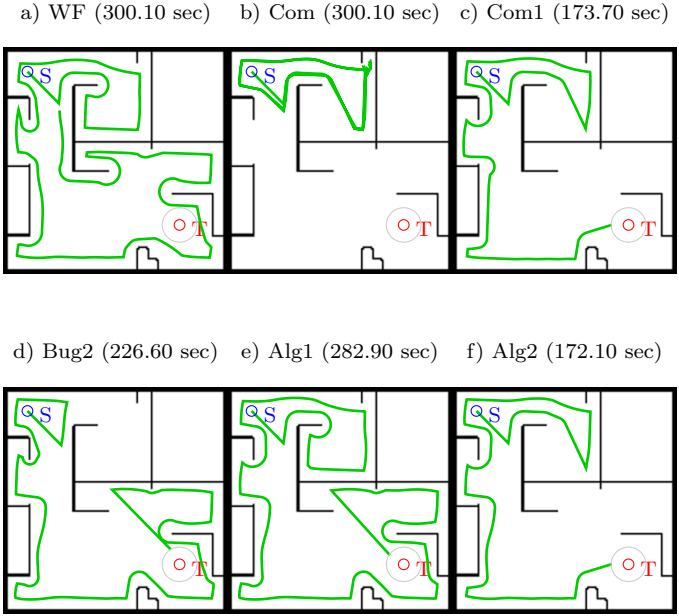}

	\caption{Behaviors of the implemented (a) wall-follower (WF) and Bug Algorithms (BAs): (b) Com, (c) Com1, (d) Bug2, (e) Alg1 and (f) Alg2   in one generated environment. The BA starts in the top left corner at Start (S) and ends withing 1 meter radius from the Target (T), with a time-limit of 300 seconds. % \#25.
		}
	\label{fig:examplenonoise}
\end{figure}

The implemented BAs' performances are now evaluated in 200 procedurally generated environments, with a constant size of 14 by 14 meters. Each BA will have one chance to navigate through the same environment with a time limit of 300 seconds. This should be a sufficient amount of time to reach the target, while preventing the simulation to run endlessly, if one of the BAs gets stuck in a loop. Each BA's success percentage is recorded, which is  the percentage of when the target is reached out of the 200 environments.  Fig.~\ref{fig:resultsbugnonoise}(a) shows the percentage of BAs that made it to the goal within the required 300 seconds, where the goal is considered reached if the BA is able to get within one meter radius.

The BA's total trajectory is recorded as well, which is normalized by dividing by an optimal path length as calculated by the A* path planning algorithm. A* will get an occupancy grid, identical to the procedurally generated environment,  which is not available to the bug algorithms by any means, and can visit all the 8 neighboring cells at each step.\footnote{An 8-connection A* will cause the path to go through a corner of an obstacle. The grid that is available A*, will include padded wall and obstacles compared to the actual environment where the Bug Algorithms will navigate in. } The normalization is applied in order to  compare the performances adequately across the generated environments, as the optimal path will be different at each iteration. Fig.~\ref{fig:resultsbugnonoise}(b) shows a box-plot of the length of the BAs' trajectories. For all BAs, all path-lengths are taken into account, including the ones that did not reach the goal. Although this can skew the statistics, a time limitation of 300 seconds will be held constant throughout the experiments to ensure consistency.

In the simulation set-up, the goal is not located near a wall, so the BA would need to leave the walls physically to reach it. Therefore, the wall-follower (WF) can should not be able to reach the goal position at all. However, there are still a few moments that the environment generator creates a situation where the WF does reaches the goal within one meter, so there is still a slim chance it is reaching the goal, as seen in Fig.~\ref{fig:resultsbugnonoise}(a). However, as WF is not moving actively towards the target, this number is marginal small. Com has more freedom in movement as it can leave the wall whenever it is free; however, is only capable to reach the goal about 60 \% of the time. Being the simplest of all the BAs, Com does not use memory, therefore, can not recognize where it has been before. Consequently, it quickly get stuck in loops, as shown in Fig.~\ref{fig:examplenonoise}(b). The last four BAs, Com1, Bug2, Alg1 and 2, have a success percentage  around the 90\% in Fig.~\ref{fig:resultsbugnonoise}(a). Still the latter two have a much shorter trajectory in comparison, which is around 2.5 times A* instead of 3.3 (Fig.~\ref{fig:resultsbugnonoise}(b)). 

In Fig.~\ref{fig:examplenonoise}(d) and (e), it can be seen that Alg1 and its ancestor Bug2 need to find the M-line first before it can leave the wall. However, this restriction seem to result in longer trajectories. The M-line-Bugs will even navigate behind the target before finding the M-line again.  Com1 and Alg2, on the other hand, will move towards the target if the chance arises, hence have more leave-opportunities along their path (Fig.~\ref{fig:examplenonoise}(c) and (f)). The outcome is that in the 200 generated environments, Com1 and Alg2 have a shorter path-length than Bug2 and Alg1 in average.\footnote{A bootstrapping based statistical similarity analysis of both the success rate and the trajectory length can be found in appendix~\ref{ap:bootbas}.}

\subsection{Experiments with Odometry Drift}\label{sec:experimentsodometrydrift}

\begin{figure}[t]
%	\centering

	%\includegraphics[width = \linewidth]{images/results_bug_trajectory_14_200_with_noise_new.png}}
\footnotesize
	\subfloat[Success percentage.]{\hspace{-0.5cm}
		\setlength\figureheight{5cm}
		\setlength\figurewidth{\linewidth}
\definecolor{mycolor1}{rgb}{0.60000,0.80000,1.00000}%
\definecolor{mycolor2}{rgb}{0.45000,0.65000,0.90000}%
\definecolor{mycolor3}{rgb}{0.30000,0.50000,0.80000}%
\definecolor{mycolor4}{rgb}{0.15000,0.35000,0.70000}%
\definecolor{mycolor5}{rgb}{0.00000,0.20000,0.60000}%
\begin{tikzpicture}

\begin{axis}[%
width=0.951\figurewidth,
height=\figureheight,
at={(0\figurewidth,0\figureheight)},
scale only axis,
bar shift auto,
xmin=0,
xmax=6,
xtick={1,2,3,4,5},
xticklabels={{Com},{Com1},{Bug2},{Alg1},{Alg2}},
ymin=0,
ymax=100,
ylabel style={font=\color{white!15!black}},
ylabel={BA reached Target [\%]},
axis background/.style={fill=white},
legend style={at={(0.03,0.97)}, anchor=north west, legend cell align=left, align=left, fill=none, draw=none},
title style={font={\small\bfseries}}, legend style={font=\tiny}, ylabel shift=-5pt 
]
\addplot[ybar, bar width=0.123, fill=mycolor1, draw=black, area legend] table[row sep=crcr] {%
1	61.5\\
2	88.5\\
3	89\\
4	88\\
5	95\\
};
\addplot[forget plot, color=white!15!black] table[row sep=crcr] {%
0	0\\
6	0\\
};
\addlegendentry{$\sigma\text{=0.00}$}

\addplot[ybar, bar width=0.123, fill=mycolor2, draw=black, area legend] table[row sep=crcr] {%
1	50.7177033492823\\
2	60.7655502392345\\
3	52.1531100478469\\
4	41.1483253588517\\
5	63.6363636363636\\
};
\addplot[forget plot, color=white!15!black] table[row sep=crcr] {%
0	0\\
6	0\\
};
\addlegendentry{$\sigma\text{=0.05}$}

\addplot[ybar, bar width=0.123, fill=mycolor3, draw=black, area legend] table[row sep=crcr] {%
1	43.0622009569378\\
2	37.3205741626794\\
3	28.2296650717703\\
4	20.5741626794258\\
5	42.5837320574163\\
};
\addplot[forget plot, color=white!15!black] table[row sep=crcr] {%
0	0\\
6	0\\
};
\addlegendentry{$\sigma\text{=0.1}$}

\addplot[ybar, bar width=0.123, fill=mycolor4, draw=black, area legend] table[row sep=crcr] {%
1	38.755980861244\\
2	33.0143540669856\\
3	15.7894736842105\\
4	14.3540669856459\\
5	30.622009569378\\
};
\addplot[forget plot, color=white!15!black] table[row sep=crcr] {%
0	0\\
6	0\\
};
\addlegendentry{$\sigma\text{=0.15}$}

\addplot[ybar, bar width=0.123, fill=mycolor5, draw=black, area legend] table[row sep=crcr] {%
1	35.8851674641148\\
2	24.8803827751196\\
3	11.0047846889952\\
4	13.8755980861244\\
5	33.4928229665072\\
};
\addplot[forget plot, color=white!15!black] table[row sep=crcr] {%
0	0\\
6	0\\
};
\addlegendentry{$\sigma\text{=0.20}$}

\end{axis}
\end{tikzpicture}%
}

   \subfloat[Normalized trajectory length]{\hspace{-0.5cm}
   	\setlength\figureheight{5cm}
   	\setlength\figurewidth{\linewidth}
   	\input{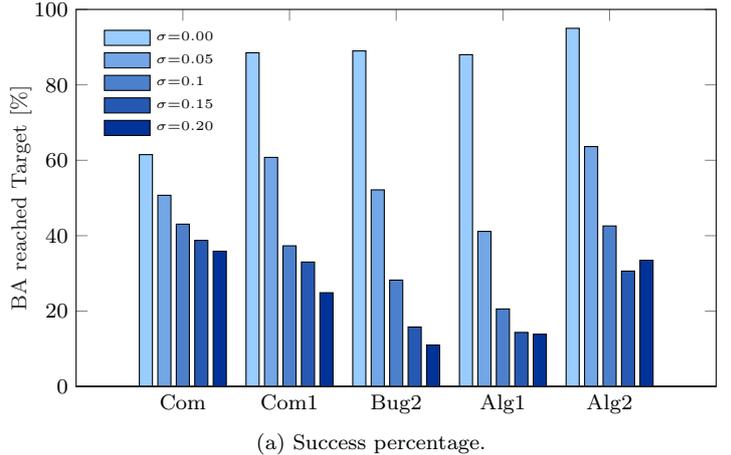}
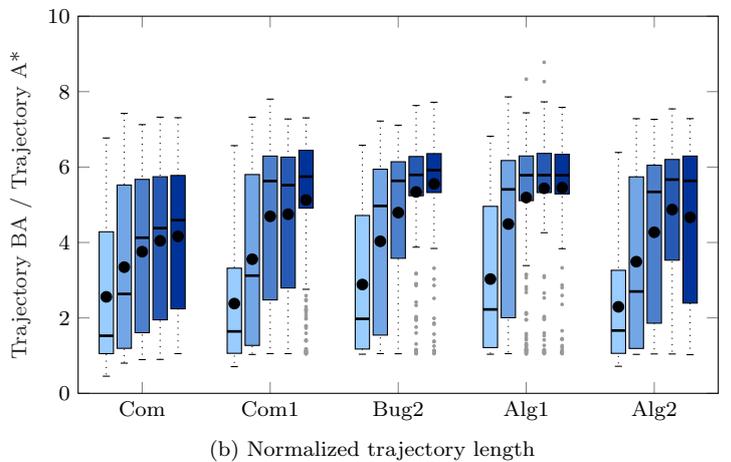

   }

	\caption{The (a) percentage of the Bug Algorithms Com, Com1, Bug2, Alg1 and Alg2, which made it to the goal of increasing velocity measurement noise ($\sigma$) which causes odometry drift, and (b) the trajectory length normalized by the optimal path calculated by A*.}
		\label{fig:resultsbugwithnoise}
	\end{figure}
	\begin{figure} [t]
		\footnotesize
		\centering
		\setlength\figureheight{7cm}
		\setlength\figurewidth{0.8\linewidth}
		\input{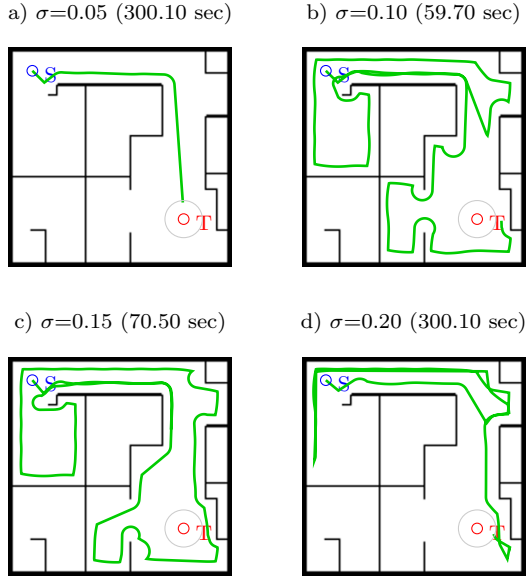}

		\caption{Example of Alg2 in environment \# 123 of the experimental testing, with increasing noise variance of $\sigma$ = (a) 0.05, (b) 0.10, (c) 0.15 and (d) 0.20. The BA starts in the top left corner at Start (S) and ends withing 1 meter radius from the Target (T), with a time-limit of 300 seconds. In (d)  Alg2 suddenly turns 180 degrees on the left side of the environment, without having seen that hit-point before.}
		\label{fig:examplealg2noise}
	\end{figure}

In this paper, we test BAs' potential for real-world navigation purposes. Therefore, we have added more realistic elements to the simulation, based on our discussion in section~\ref{sec:BArealworld}.  In the absence of an exact global position, BAs will  need to rely on odometry. Therefore, this section will investigate the effects of odometry drift. We assume that the BA will know its own and the target's position at the start of the experiments, but it   has to keep them up-to-date with its own, noisy, velocity measurements.  For these experiments, we assume that the position estimate is acquired by the latter assumption, namely:

\begin{equation}
\mathbf{\tilde x}_t = \mathbf{ \tilde x}_{t-1} + \mathbf{\dot{ \tilde{ x}}}_{t-1}
\end{equation}
, where $\mathbf{ \mathbf{\tilde x_t}}$ is the x- and y-position estimate at a given time. 
$ \mathbf{\dot{ \tilde{ x}}}_{t-1}$ is assumed to be $\mathcal{N}(\mathbf{u}_{t-1},\sigma)$, where $\mathbf{u}$ is the actual velocity, from which the outcome on the system consists of noise with a standard deviation of $\sigma$.

Fig.~\ref{fig:resultsbugwithnoise} shows the impact on the performances of the BAs when exposed to odometry drift due to noisy velocity estimates, with a $\sigma$ of 0.05, 0.10, 0.15 and 0.2.
In Fig.~\ref{fig:resultsbugwithnoise}(a) indicates a significant drop in all the BAs' success percentage with an increasing $\sigma$.
In Fig.~\ref{fig:resultsbugwithnoise}(b) we see that it has a large effect on the trajectory length overall, although there is a less significant degeneration of the Angle-Bugs' performance (Com, Com1 and Alg2). Bug2's and Alg2's performances took the deepest dive with a relatively small increment of the odometry drift, whereas Com's performance only gradually decreased. As Com does not save any position or distance-to-goal at hit-points, only its bearing estimate towards the goal is effected by faulty velocity estimates, resulting in the simplest BA outperforming the rest with $\sigma>0.05$.  Alg2 already lost its advantage to recognize previously visited places, as its success-rate is similar, if not lower, than Com1 at a $\sigma$ of 0.2. 

However, both Alg1 and Alg2 show signs of stagnation from $\sigma$ = 0.15 and on, as their performances does not seem to decrease any further and even seem to improve slightly. At that point, it could be that it would accidentally recognize a previously hit-point at a location where
 it has not been before due to the odometry drift. Although seemingly unwanted,  this randomness could have helped the BA to get out of difficult situations, as in Fig.~\ref{fig:examplealg2noise} with Alg2. Although it is still successful at a $\sigma$ = 0.05 (Fig.~\ref{fig:examplealg2noise}(a)), at a velocity measurement noise of  $\sigma> 0.05$, Alg2 is already unable to go straight towards the goal in Fig~\ref{fig:examplealg2noise}(b-c), prematurely hitting a wall and navigating backwards. Fig.~\ref{fig:examplealg2noise}(d) shows that at a $\sigma$ = 0.2, Alg2 suddenly encounters a place that it thinks it has been before and turns around during wall following. Although the BA's observation is false, it did put Alg2 back into a situation where it could reach the goal once again, even though it needed a longer trajectory than without odometry drift.\footnote{Statistical correlation analysis of the effect of the increasing odometery noise both the success rate and trajectory length can be found in appendix~\ref{ap:corrodometry}.}

\subsection{Experiments with False Positive and False Negative Recognition Rate}\label{sec:experimentsFPFN}
\begin{figure} [t]
	\centering
	
	\subfloat[Trajectory length (FP)]{
		\hspace{-0.5cm}
		\footnotesize
		\setlength\figureheight{6cm}
		\setlength\figurewidth{0.38\linewidth}
		\input{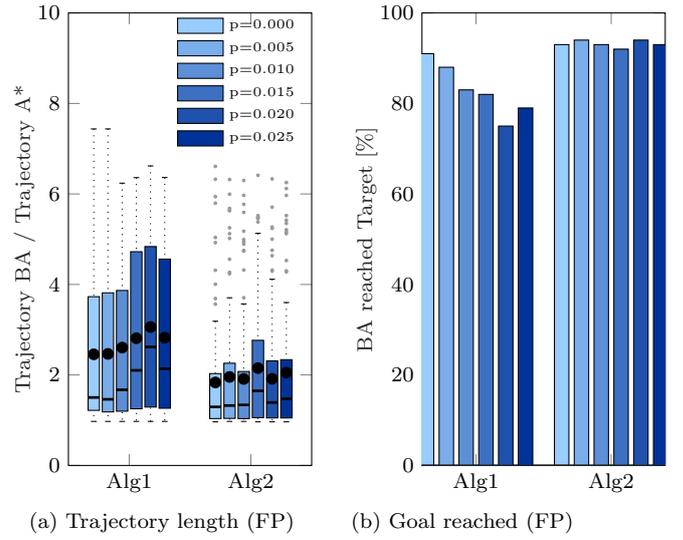}}\hspace{-2.2cm}\captionsetup[subfigure]{oneside,margin={1.5cm,0cm}}\subfloat[Goal reached (FP)]{		\footnotesize\setlength\figureheight{6cm}
		\setlength\figurewidth{0.38\linewidth}
\definecolor{mycolor1}{rgb}{0.60000,0.80000,1.00000}%
\definecolor{mycolor2}{rgb}{0.48000,0.68000,0.92000}%
\definecolor{mycolor3}{rgb}{0.36000,0.56000,0.84000}%
\definecolor{mycolor4}{rgb}{0.24000,0.44000,0.76000}%
\definecolor{mycolor5}{rgb}{0.12000,0.32000,0.68000}%
\definecolor{mycolor6}{rgb}{0.00000,0.20000,0.60000}%
\begin{tikzpicture}

\begin{axis}[%
width=0.951\figurewidth,
height=\figureheight,
at={(0\figurewidth,0\figureheight)},
scale only axis,
bar shift auto,
xmin=0.6,
xmax=2.4,
xtick={1,2},
xticklabels={{Alg1},{Alg2}},
ymin=0,
ymax=100,
ylabel style={font=\color{white!15!black}},
ylabel={BA reached Target [\%]},
axis background/.style={fill=white},
legend style={at={(-1.042,0.685)}, anchor=south west, legend cell align=left, align=left, fill=none, draw=none},
title style={font={\small\bfseries}}, legend style={font=\tiny}, ylabel shift=-5pt
]
\addplot[ybar, bar width=0.107, fill=mycolor1, draw=black, area legend] table[row sep=crcr] {%
1	91\\
2	93\\
};
\addplot[forget plot, color=white!15!black] table[row sep=crcr] {%
0.6	0\\
2.4	0\\
};
\addlegendentry{p=0.000}

\addplot[ybar, bar width=0.107, fill=mycolor2, draw=black, area legend] table[row sep=crcr] {%
1	88\\
2	94\\
};
\addplot[forget plot, color=white!15!black] table[row sep=crcr] {%
0.6	0\\
2.4	0\\
};
\addlegendentry{p=0.005}

\addplot[ybar, bar width=0.107, fill=mycolor3, draw=black, area legend] table[row sep=crcr] {%
1	83\\
2	93\\
};
\addplot[forget plot, color=white!15!black] table[row sep=crcr] {%
0.6	0\\
2.4	0\\
};
\addlegendentry{p=0.010}

\addplot[ybar, bar width=0.107, fill=mycolor4, draw=black, area legend] table[row sep=crcr] {%
1	82\\
2	92\\
};
\addplot[forget plot, color=white!15!black] table[row sep=crcr] {%
0.6	0\\
2.4	0\\
};
\addlegendentry{p=0.015}

\addplot[ybar, bar width=0.107, fill=mycolor5, draw=black, area legend] table[row sep=crcr] {%
1	75\\
2	94\\
};
\addplot[forget plot, color=white!15!black] table[row sep=crcr] {%
0.6	0\\
2.4	0\\
};
\addlegendentry{p=0.020}

\addplot[ybar, bar width=0.107, fill=mycolor6, draw=black, area legend] table[row sep=crcr] {%
1	79\\
2	93\\
};
\addplot[forget plot, color=white!15!black] table[row sep=crcr] {%
0.6	0\\
2.4	0\\
};
\addlegendentry{p=0.025}

\end{axis}
\end{tikzpicture}%
}	
	
			\hspace{-1.5cm}
	\subfloat[Trajectory length (FN)]{
			\hspace{1cm}
		\footnotesize
		\setlength\figureheight{6cm}
		\setlength\figurewidth{0.38\linewidth}
		\input{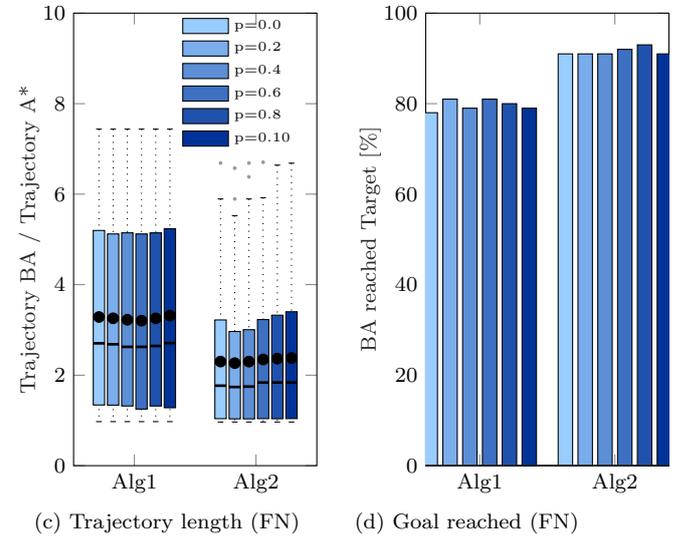}}\hspace{-2.2cm}\captionsetup[subfigure]{oneside,margin={1.5cm,0cm}}\subfloat[Goal reached (FN)]{		\footnotesize\setlength\figureheight{6cm}
		\setlength\figurewidth{0.38\linewidth}
\definecolor{mycolor1}{rgb}{0.60000,0.80000,1.00000}%
\definecolor{mycolor2}{rgb}{0.48000,0.68000,0.92000}%
\definecolor{mycolor3}{rgb}{0.36000,0.56000,0.84000}%
\definecolor{mycolor4}{rgb}{0.24000,0.44000,0.76000}%
\definecolor{mycolor5}{rgb}{0.12000,0.32000,0.68000}%
\definecolor{mycolor6}{rgb}{0.00000,0.20000,0.60000}%
\begin{tikzpicture}

\begin{axis}[%
width=0.951\figurewidth,
height=\figureheight,
at={(0\figurewidth,0\figureheight)},
scale only axis,
bar shift auto,
xmin=0.6,
xmax=2.4,
xtick={1,2},
xticklabels={{Alg1},{Alg2}},
ymin=0,
ymax=100,
ylabel style={font=\color{white!15!black}},
ylabel={BA reached Target [\%]},
axis background/.style={fill=white},
legend style={at={(-1.036,0.685)}, anchor=south west, legend cell align=left, align=left, fill=none, draw=none},
title style={font={\small\bfseries}}, legend style={font=\tiny}, ylabel shift=-5pt
]
\addplot[ybar, bar width=0.107, fill=mycolor1, draw=black, area legend] table[row sep=crcr] {%
1	78\\
2	91\\
};
\addplot[forget plot, color=white!15!black] table[row sep=crcr] {%
0.6	0\\
2.4	0\\
};
\addlegendentry{p=0.0}

\addplot[ybar, bar width=0.107, fill=mycolor2, draw=black, area legend] table[row sep=crcr] {%
1	81\\
2	91\\
};
\addplot[forget plot, color=white!15!black] table[row sep=crcr] {%
0.6	0\\
2.4	0\\
};
\addlegendentry{p=0.2}

\addplot[ybar, bar width=0.107, fill=mycolor3, draw=black, area legend] table[row sep=crcr] {%
1	79\\
2	91\\
};
\addplot[forget plot, color=white!15!black] table[row sep=crcr] {%
0.6	0\\
2.4	0\\
};
\addlegendentry{p=0.4}

\addplot[ybar, bar width=0.107, fill=mycolor4, draw=black, area legend] table[row sep=crcr] {%
1	81\\
2	92\\
};
\addplot[forget plot, color=white!15!black] table[row sep=crcr] {%
0.6	0\\
2.4	0\\
};
\addlegendentry{p=0.6}

\addplot[ybar, bar width=0.107, fill=mycolor5, draw=black, area legend] table[row sep=crcr] {%
1	80\\
2	93\\
};
\addplot[forget plot, color=white!15!black] table[row sep=crcr] {%
0.6	0\\
2.4	0\\
};
\addlegendentry{p=0.8}

\addplot[ybar, bar width=0.107, fill=mycolor6, draw=black, area legend] table[row sep=crcr] {%
1	79\\
2	91\\
};
\addplot[forget plot, color=white!15!black] table[row sep=crcr] {%
0.6	0\\
2.4	0\\
};
\addlegendentry{p=0.10}

\end{axis}
\end{tikzpicture}%
}

	%	\subfloat[Trajectory length (FN)]{\includegraphics[width = 0.49\linewidth]{images/results_bug_trajectory_14_100_with_FN.png}}\hfill\subfloat[Goal reached (FN)]{\includegraphics[width = 0.49\linewidth]{images/results_bug_percentage_14_100_with_FN.png}}
	
	\caption{The (a\&b) measured trajectory length and (c\&d) percentage of Alg1 and Alg2 reaching the goal, with a varying chance ($p$) of a false-positive (FP) or a false-negative (FN) of a previous recognized point to occur. }
	\label{fig:resultsbugwithFP}
\end{figure}
\begin{figure} [t]

	\footnotesize\setlength\figureheight{9cm}
	\setlength\figurewidth{\linewidth}
	\input{result_plots/results_bug_odometry_noise_env_50_alg1.tex}
	\caption{An example environment with the trajectories of Alg1 with increasing chance($p$) of False Positives (FP) to occur, with $p$(FP) = (a) 0.0, (b) 0.2, (c) 0.4, (d) 0.6, (e) 0.8 and (f) 1.0.The BA starts in the top left corner at Start (S) and ends withing 1 meter radius from the Target (T), with a time-limit of 300 seconds. }%#50
	\label{fig:resultsexampleFP}
\end{figure}

BAs can also recognize previous hit-points based on scene recognitions. In this paper, we will not use the techniques and descriptors discussed in subsection~\ref{sec:BArealworld}, but will simulate their performance through false-negative (FN) and false-positive (FP) recall rates. With an increasing probability ($p$) of a uniform distribution, the chances of a previously visited hit-point being falsely recognized at a different location (FP)  or not being recognized at the right position (FN) will increase. 

Of the implemented BA, only Alg1 and Alg2 specifically use previously visited locations to change their local wall-following direction from right- to left- sided. In Fig.~\ref{fig:resultsbugwithFP}, they are being evaluated with an increasing $p$(FP) in Fig.~\ref{fig:resultsbugwithFP}(a\&b) or $p$(FN) in Fig.~\ref{fig:resultsbugwithFP}(c\&d) over 100 generated environments. At a $p$(FP)=0.005, there is a chance of FP occurring 1-2 times (0.5 \%) during the run-time of 300 second and at $p$(FP)=0.025 a chance of 7-8 times (2.5 \%). At $p$(FN) = 0.2, every time the BA encounters a previous hit-point, there is a 20 \% chance that it will not recognize it and at $p$(FN)=1.0, the hit-point will never be recalled.

Fig.~\ref{fig:resultsbugwithFP}(a) and (b) shows that increasing the $p$(FP) has more effect on the performance of Alg1 than Alg2. An example of Alg1's behavior in an environment with an increasing $p$(FP) is shown in Fig.~\ref{fig:resultsexampleFP}(a). From $p$(FP)=0.2 and on, Alg1 misses the sparse and crucial places on the M-line where it needs to leave the wall, at the moments it prematurely detects a hit-point (\ref{fig:resultsexampleFP}(b-f)). Alg2 has fewer leave-restrictions and can move towards the target whenever its path is clear. Thanks to this  flexible behavior, it seems to be less sensitive to more frequent occurrences of FP. 

In Fig.~\ref{fig:resultsbugwithFP}(c) and (d), the effects of a higher FN rate is shown; however, both Alg1 as Alg2 seemed to be hardly effected by it. The only trend that could be noticed is for Alg2 as the variance of the trajectory length slowly creeps up with an increasing $p$(FN) in Fig.~\ref{fig:resultsbugwithFP}(c). When $p$(FN) = 1.0, then Alg2 is completely identical to the implemented Com1, as it only remembers the range measurements at hit-points as a  leave-condition. The same goes for Alg1, which transforms into its ancestor Bug2 with $p$(FN). As both Bug2 and Com1 have the ability to get out of a loop,  almost no difference can be noticed in the success rate of Fig.~\ref{fig:resultsbugwithFP}(d) with $p$(FN) = 0.0 and 1.0 for both Alg1 and Alg2.\footnote{Statistical correlation analysis of the effect of the increasing recognition failure rate on both the success rate and trajectory length can be found in appendix~\ref{ap:corrFNFP}.}

\subsection{Experiments with Distance Measurement Noise}\label{sec:experimentsrangenoise}
	
	\begin{figure} [t]
			\subfloat[Trajectory length]{
				\hspace{-0.5cm}
				\footnotesize
				\setlength\figureheight{6cm}
				\setlength\figurewidth{0.38\linewidth}
				\input{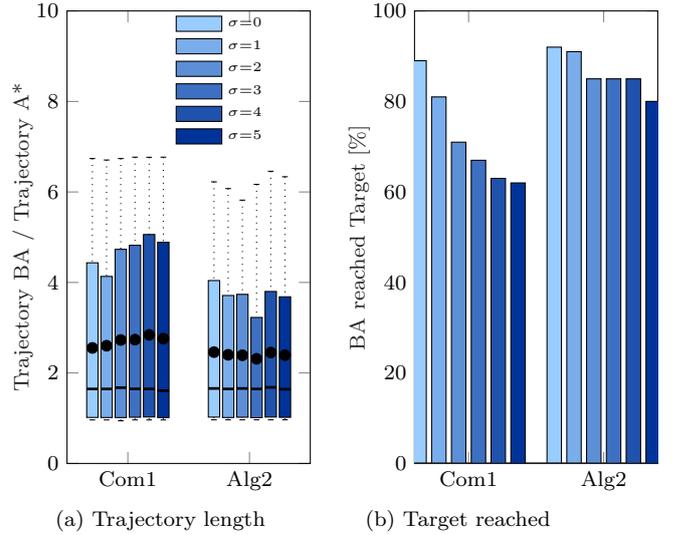}}\hspace{-2.2cm}\captionsetup[subfigure]{oneside,margin={1.5cm,0cm}}\subfloat[Target reached]{		\footnotesize\setlength\figureheight{6cm}
				\setlength\figurewidth{0.38\linewidth}
\definecolor{mycolor1}{rgb}{0.60000,0.80000,1.00000}%
\definecolor{mycolor2}{rgb}{0.48000,0.68000,0.92000}%
\definecolor{mycolor3}{rgb}{0.36000,0.56000,0.84000}%
\definecolor{mycolor4}{rgb}{0.24000,0.44000,0.76000}%
\definecolor{mycolor5}{rgb}{0.12000,0.32000,0.68000}%
\definecolor{mycolor6}{rgb}{0.00000,0.20000,0.60000}%
\begin{tikzpicture}

\begin{axis}[%
width=0.951\figurewidth,
height=\figureheight,
at={(0\figurewidth,0\figureheight)},
scale only axis,
bar shift auto,
xmin=0.6,
xmax=2.4,
xtick={1,2},
xticklabels={{Com1},{Alg2}},
ymin=0,
ymax=100,
ylabel style={font=\color{white!15!black}},
ylabel={BA reached Target [\%]},
axis background/.style={fill=white},
legend style={at={(-1.017,0.685)}, anchor=south west, legend cell align=left, align=left, fill=none, draw=none},
title style={font={\small\bfseries}}, legend style={font=\tiny}, ylabel shift=-5pt
]
\addplot[ybar, bar width=0.107, fill=mycolor1, draw=black, area legend] table[row sep=crcr] {%
1	89\\
2	92\\
};
\addplot[forget plot, color=white!15!black] table[row sep=crcr] {%
0.6	0\\
2.4	0\\
};
\addlegendentry{$\sigma\text{=0}$}

\addplot[ybar, bar width=0.107, fill=mycolor2, draw=black, area legend] table[row sep=crcr] {%
1	81\\
2	91\\
};
\addplot[forget plot, color=white!15!black] table[row sep=crcr] {%
0.6	0\\
2.4	0\\
};
\addlegendentry{$\sigma\text{=1}$}

\addplot[ybar, bar width=0.107, fill=mycolor3, draw=black, area legend] table[row sep=crcr] {%
1	71\\
2	85\\
};
\addplot[forget plot, color=white!15!black] table[row sep=crcr] {%
0.6	0\\
2.4	0\\
};
\addlegendentry{$\sigma\text{=2}$}

\addplot[ybar, bar width=0.107, fill=mycolor4, draw=black, area legend] table[row sep=crcr] {%
1	67\\
2	85\\
};
\addplot[forget plot, color=white!15!black] table[row sep=crcr] {%
0.6	0\\
2.4	0\\
};
\addlegendentry{$\sigma\text{=3}$}

\addplot[ybar, bar width=0.107, fill=mycolor5, draw=black, area legend] table[row sep=crcr] {%
1	63\\
2	85\\
};
\addplot[forget plot, color=white!15!black] table[row sep=crcr] {%
0.6	0\\
2.4	0\\
};
\addlegendentry{$\sigma\text{=4}$}

\addplot[ybar, bar width=0.107, fill=mycolor6, draw=black, area legend] table[row sep=crcr] {%
1	62\\
2	80\\
};
\addplot[forget plot, color=white!15!black] table[row sep=crcr] {%
0.6	0\\
2.4	0\\
};
\addlegendentry{$\sigma\text{=5}$}

\end{axis}
\end{tikzpicture}%
}	
	%	\centeringrange_noise
		%\subfloat[]{\includegraphics[width = 0.49\linewidth]{images/results_bug_trajectory_14_100_with_range_noise.png}}\hfill\subfloat[]{\includegraphics[width = 0.49\linewidth]{images/results_bug_percentage_14_100_with_range_noise.png}}	
		\caption{The performance of Alg2 and IBug with varying distance measurement noise ($\sigma$) in meters, in the (a) normalized trajectory length  and percentage (b) of bugs who made it to the goal.}
		\label{fig:resultsbugwithrangenoise}
	\end{figure}
\begin{figure} [t]
	
	\footnotesize\setlength\figureheight{9cm}
	\setlength\figurewidth{\linewidth}
	\input{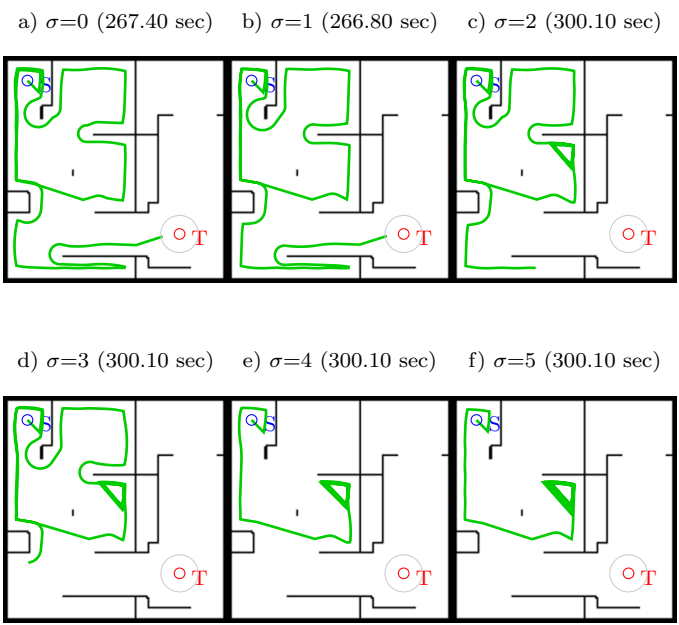}
	%\centerline{
	%\includegraphics[width = 1.2\linewidth]{images/example_noise_range_env_88_Ibug.png}}
	\caption{An example environment with the trajectories of Com1 with increasing distance measurement noise variance ($\sigma$) in meters of (a) 0, (b) 1, (c) 2, (d) 3, (e) 4 and (f) 5 meters. The BA starts in the top left corner at Start (S) and ends within 1 meter radius from the Target (T), with a time-limit of 300 seconds.}%88
	\label{fig:resultsexamplerangenoise}
\end{figure}

%BAs could also  use  a Distance-to-Target (DT)  measurement, so  we assume that the agents are carrying a sensor able to determine the DT, such as received signal strength intensity of BlueTooth (\cite{bargh2008indoor}) or Ultra-Wide Band (UWB, \cite{guo2017ultra}). However, none of these sensors are perfect. DT estimation by RSSI by BlueTooth could get an error up to 2 meters and can not practically determine a range from 5 meters on\footnote{This is based on a Bled112 Bluegiga Bluetooth module} (\cite{coppola2016board}), which is highly determined by the amount of obstacle clutter. UWB has better statistics, with a standard deviation of 0.2 meters  and a maximum limit up to 100 meters\footnote{This is based on a DecaWave UWB module in ranging mode}, without being interfered much by walls and obstacles in between. Even so, the high energy expenditure of the latter could be a valid reason to prefer the energy efficient BlueTooth dongle. 

BAs could also  use  a Distance-to-Target (DT)  measurement, so here we assume that the agents are carrying a sensor able to determine this. Com1 and Alg2 both save previous DT measurements to prevent getting stuck in a loop in some situations. In Fig.~\ref{fig:resultsbugwithrangenoise}, we are showing the (a) trajectory length and (b) success rate of the increasing standard deviation of the DT noise, while keeping both the velocity measurement noise (odometry drift) and the FP \& FN rate at 0.0.  The noisy DT measurements ($\tilde d(x,T)_t$) at time $t$ are modeled by ${\tilde d(x,T)_t} = \mathcal{N}(d(x,T)_t,\sigma) $, where $d(x,T)_t$ is a scalar that stands for the true DT at time $t$ and $\sigma$ is the standard deviation of the noise.  The degrading performance in both trajectory length and success percentage for increasing $\sigma$ is more noticeable for Com1 than for Alg2. Com1's only mechanism to get out of a potential loop is to compare its current DT with a  saved one to decide when to leave the wall. Once it is gradually losing this capability with the noisier DT measurements, its behavior will become more and more similar to Com's, as observed in Fig.~\ref{fig:resultsexamplerangenoise}. Moreover, Com1's success percentages in Fig.~\ref{fig:resultsexamplerangenoise}(b) drops to around 60 percent at a $\sigma$=6 meters, which is equivalent to Com's score in Fig.~\ref{fig:resultsbugnonoise}(b). Alg2 is less affected by the increasing DT noise, which is likely because it can rely on memorized position as an additional leave condition.\footnote{Statistical correlation analysis of the effect of the increasing DT measurement noise on both the success rate and trajectory length can be found in appendix~\ref{ap:corrrange}.}

\section{Discussion}\label{ch:discussion}

This section will reflect on both the experimental set-up and results. The modeled real-world conditions will be discussed first, including the implementation details of the simulation and the chosen noise-models and Bug Algorithms (BAs). Here we will give some suggestions for future development in this topic. Afterwards, we will discuss the results from our experiments, from which we will determine which BA aspects work or do not work for real-world scenarios. This discussion will be concluded on in the last section of this comparative study.

\subsection{Modeling Real-World Conditions}

 BAs are a seemingly ideal indoor navigation paradigm for tiny robotic platforms with limited recourses. Potentially, they could only take up a small fraction of the on-board computer's capacity, which opens up space for other computations and tasks. Although the paths generated are sub-optimal compared to path-planning algorithms as A*, no map is needed or needs to be generated. With the target's location in mind, the BA reacts locally on obstacles and only saves small bits of information in order to converge. Nevertheless, we  established that the BAs, presented in section~\ref{ch:bugalgorithms}, tend to over-rely on a perfect localization, which can not be guaranteed for indoor environments. 

 If no global localization scheme can be set-up, the BA needs to rely on its noisy on-board sensors to know where it is and integrate this knowledge with the target's position. In section~\ref{ch:implemetation}, we reflected on several issues that a robotic implementation of a BA will come across. This includes: an increasing odometry drift, a mis-match between its measured position and the ground truth; recognition failures, i.e. when it fails to recognize a previous location or falsely detects one; and noisy distance to target measurements, which could interfere with the suitable leaving-condition. There are other sensor-noise and failures to consider, such as the noise in the laser-range sensors or (stereo-)cameras for (local) boundary/wall following. However, in this paper, we focused on the global position estimation instead, as this is an issue that all bug algorithms have to deal with and is less dependent on the implemented platform. Moreover, we aimed to keep the wall-following behavior identical among the implemented BAs.

In the experimental setup, section~\ref{ch:experimentalsetup}, we selected a set of suitable BAs to experiment on and motivated that choice. We believe that this selection represented most issues of real-world implementation well enough to draw generic conclusions, applicable to the more current BA variants, such as TangentBug (\cite{Kamon1998}). However, for future work, we could also look at newer BAs, where we could include the earlier mentioned local sensor noise. Next to this, the ARGoS simulator and environment generator was very useful for this paper's experiments, as it was able to generate new environments at a high pace and run the experiments quicker than real-time. This enabled us to test the BAs on hundreds of environments, leading to more reliable results. Nevertheless, further development of these experiments must be performed in a more realistic simulation, with more types of obstacles and visual representations, to induce more challenges of a typical indoor navigation task. 

For the experiments themselves in, section~\ref{ch:experimentresults}, we used simple noise models, i.e. using a Gaussian probability distribution for the odometry drift and noisy range measurements, and a pseudo-random number generator for FNs and FPs occurrences. Future work could look at more realistic noise characteristics. For ground-bound robots, for instance, wheel slippage is determined by the materials used and the friction with the floor. If visual-odometry is used, Gaussian noise could very well be applied, however the texture of the environment is crucial to the variance. The FP \& FN occurrences are also very much determined by the features of the environment, as aliasing could occur at areas that are very similar. There is no equal probability of these failures to happen throughout the trajectory of the bug. Moreover, distance measurements by radio beacons not only suffer from regular noise around the mean, but have to endure a whole range of disturbances. This includes uneven directional propagation noise, the reflection off the walls, interference of other signals. For the experiments in this paper, we wanted to have more control over the noisy measurements to find a clear correlation between its severity and the performance of the BAs, so we restricted ourselves to use the basic versions of the noise models. However, these considerations  be included for an even more realistic simulation in future work.

\subsection{Experimental Results}

Generally, our experiments showed that all BAs performed worse with a higher odometry drift, noisier range measurements and increasing failure cases. The most noticeable feature, is that the BAs did not all have a similar drop in performances, which is especially noticeable with increasing odometry drift in Fig.~\ref{fig:resultsbugwithnoise}. Some had a more severe response than others, namely those using memory. Com, being the simplest of all BAs, started out as the worst one of the six, to the best performing with only standard deviation of 0.1 m/s in the velocity estimation. As it only uses the odometry to get a range and bearing to the goal and nothing else, there are less "bad" decisions it could make based on it. Since odometry is likely to be noisy on very small robots, such simplicity may be the better strategy. Nevertheless, although Com is less influenced by odometry drift, it success rate still drops to 40\%, which is still a low score. In general, it is ill-advised to have BAs solely rely on odometry alone.

In section~\ref{sec:experimentsFPFN} and \ref{sec:experimentsrangenoise}, we also assumed that the BAs will also have access to  measurements other than odometry. Although a decrease in performance was noticed in all the tested BAs with these specific features (Com1, Alg1 and Alg2), it became evident that Alg2 is the most resilient algorithm.  With increasing FN \& FP occurrences, Alg1's performance was noticeably decreasing but Alg2 was hardly affected. This indicates that the M-line-Bugs, as Alg1, seem to have a disadvantage over Angle-Bugs, as Alg1, due to their restrictive leave-condition. This is also noticeable in section~\ref{sec:experimentsodometrydrift}, as M-line-Bugs suffered the most from increasing odometry noise. If real-world conditions apply, BAs should rather be able to leave the wall/obstacle whenever there is the possibility to do so.

The same goes for noisy distance-to-target measurements (section~\ref{sec:experimentsrangenoise}), where Com1  is performing worse than Alg2. The reasoning behind this observation is  simple: Alg2 is using more mechanisms to get out of complex situations, namely remembering range measurements and locations of previous hit-points. If one of these  mechanisms perform badly, then Alg2 can fall back on the other one. Now these measures are operating separately and have a different behavioral outcome; however, it could be more beneficial to a BA if they were fused together or used for cross validation and checking if the bug is stuck in a loop. Nevertheless, it is of great interest to have multiple types of measurements to rely on, either concerning position of the robot itself or the relative position of the goal.

\section{Conclusion}\label{sc:conclusion}

This paper investigates the potential of Bug Algorithms as a computationally efficient method for robotic navigation. Although the general idea behind the methods seems ideal for implementation of light-weight robots, the literature survey shows that many of their variants rely on either a global localization system or perfect on-board sensors. Our simulation experiments evaluated several implemented Bug Algorithms with varying noisy measurements and failure cases, which showed a significant performance degradation of all algorithms. This indicates that Bug Algorithms can not simply be implemented as they are on a navigating robot, which has to rely on only on-board sensors without any external help. The experimental results did, however, shed some light on how these techniques can be enhanced. Simplicity is a key element, as the most basic Bug Algorithm, Com, was also the one that was the most resilient to odometry drift. Another crucial element is a robust loop detection system, where the robot should not just rely on one but on multiple  measured variables, especially in realistic, noise-inducing, environments. Considering these observations in the design of new Bug Algorithms, will make them suitable for the autonomous navigation of tiny robotic platform with limited computational recourses.

\section*{Acknowledgements}
This work has been funded by the NWO grant of Natural Intelligence. The research has been conducted at the Delft University of Technology, Faculty of AeroSpace Engineering, Department of Control and Simulation and Liverpool University at the Faculty of Computer Science, SmartLab. I would like to thank James Butterworth for helping me with setting up the ARGoS simulation and brainstorming about Bug-Algorithms in general.

\section*{References}
\bibliographystyle{elsarticle-harv} 

\bibliography{library_RAAS_2018}

\appendix

\section*{Appendices}
\renewcommand{\thesection}{A}
\section{Pseudo-Code Bug Algorithms}\label{ap:pseudoBAs}
The pseudo-code for Com, Com1, Bug2, Alg1 and Alg2, is listed in Algorithm~\ref{alg:com}, ~\ref{alg:com1}, ~\ref{alg:bug2}, ~\ref{alg:alg1} and ~\ref{alg:alg2} respectively. T stands for target and $s_{WF}$ is a variable that determines if the wall-following is right- ($s_{WF}$=1) or left-sided ($s_{WF}$=-1). $r_{local}$ stand for local sensor measurements, which can be either contact- or range-sensors. $x_global$ stands for the global position estimate of the Bug Algorithm. $d(H,T)_{prev}$ stands for the previous distance of the hit-point to target and $d(x_{global},T)$ stands for the current distance from BA to target. $list_{hp}$ stand for a list of previously encountered hit-points. $v$ is the control output for the forward velocity of the robot and $c_v$ is the fixed velocity constant. $\omega$ is the control output for the heading of the robot in rad/s and $c_\omega$ is a fixed rate constant, to control the speed of the robot's turns.

\begin{algorithm}[H]
		\footnotesize
	\caption{The pseudo-code for the state-machine of Com. }\label{alg:com}
	\begin{algorithmic}
		
		\State Init:{ $state$ = "forward", $s_{WF}$}
		\Require{$c_v$, $c_\omega, x_{global} r_{local}, list_{hp}$}

		\Function{Com}{}
			\If{$state$ is "forward" }
				\State{$v \gets c_v$}
				\State{$\omega \gets 0$ } 
				\If{Obstacle is hit}
					\State {$state \gets$ "wall\_following"}
				\EndIf
			\ElsIf{$state$ is "wall\_following"}
				\State{ $[v,~ \omega]\gets$  \texttt{Wall\_Following($c_v, c_\omega,s_{WF} , r_{local}$)}}
				\Comment{See \ref{ap:wallfollowing}}
				\If{Way towards T is free} \Comment{Based on $r_{local}$}
					\State $state \gets$ "rotate\_to\_target"
				\EndIf
			\ElsIf{$state$ is "rotate\_to\_target"}
				\State{$v \gets 0$} 
				\State{$\omega \gets c_\omega$}
				\If{Heading BA same as direction T}
					\State{$state \gets$ "forward"}
				\EndIf	
			\EndIf	
			\Return {$v, \omega$}
		\EndFunction

	\end{algorithmic}
\end{algorithm}

\begin{algorithm}[H]
		\footnotesize
	\caption{The pseudo-code for  the state-machine of Com1. }\label{alg:com1}
	\begin{algorithmic}
		
		\State Init:{ $state$ = "forward", $s_{WF}=1$}
		\Require{$c_v$, $c_\omega, r_{local}$}
		
		\Function{Com}{}
		\If{$state$ is "forward" }
		\State{$v \gets c_v$} 
		\State{$\omega \gets 0$ } 
		\If{Obstacle is hit}
		\State {$d(H,T) \gets d(x_{global},T)$}
		\State {$state \gets$ "wall\_following"}
		\EndIf
		\ElsIf{$state$ is "wall\_following"}
		\State{ $[v,~ \omega]\gets$  \texttt{Wall\_Following($c_v, c_\omega,s_{WF} , r_{local}$)}}
		\Comment{See \ref{ap:wallfollowing}}
		\If{Way towards T is free and  $d(x_{global},T)$<$d(H,T)$} 
		\State $state \gets$ "rotate\_to\_target"
		\EndIf
		\ElsIf{$state$ is "rotate\_to\_target"}
		\State{$v \gets 0$} 
		\State{$\omega \gets c_\omega$}
		\If{Heading BA same as direction T}
		\State{$state \gets$ "forward"}
		\EndIf	
		\EndIf	
		\Return {$v, \omega$}
		\EndFunction

	\end{algorithmic}
\end{algorithm}

\begin{algorithm}[H]
		\footnotesize
	\caption{The pseudo-code for the state-machine of Bug2. }\label{alg:bug2}
	\begin{algorithmic}
		
		\State Init:{ $state$ = "forward", $s_{WF}=1$}
		\Require{$M-line$,$c_v$, $c_\omega, x_{global} r_{local}$}
		
		\Function{Com}{}
		\If{$state$ is "forward" }
		\State{$v \gets c_v$}
		\State{$\omega \gets 0$ } 
		\If{Obstacle is hit}
		\State {$state \gets$ "wall\_following"}
		\EndIf
		\ElsIf{$state$ is "wall\_following"}
		\State{ $[v,~ \omega]\gets$  \texttt{Wall\_Following($c_v, c_\omega,s_{WF} , r_{local}$)}}
		\Comment{See \ref{ap:wallfollowing}}
		\If{$M-line$ is hit and BA is closer to T} 
		\State $state \gets$ "rotate\_to\_target"
		\EndIf
		\ElsIf{$state$ is "rotate\_to\_target"}
		\State{$v \gets 0$} 
		\State{$\omega \gets c_\omega$}
		\If{Heading BA same as direction T}
		\State{$state \gets$ "forward"}
		\EndIf	
		\EndIf	
		\Return {$v, \omega$}
		\EndFunction

	\end{algorithmic}
\end{algorithm}
\begin{algorithm}[h]
	\footnotesize
	\caption{The pseudo-code for the state-machine of Alg1. }\label{alg:alg1}
	\begin{algorithmic}
		
		\State Init:{ $state$ = "forward", $s_{WF}=1$, , $list_{HP}$=[~]}
		\Require{$M-line$,$c_v$, $c_\omega, x_{global} r_{local}$}
		
		\Function{Com}{}
		\If{$state$ is "forward" }
		\State{$v \gets c_v$}
		\State{$\omega \gets 0$ } 
		\If{Obstacle is hit}
		\State {$list_{HP}\gets[list_{HP}, x_{global}]$}
		\State {$state \gets$ "wall\_following"}
		\EndIf
		\ElsIf{$state$ is "wall\_following"}
		\State{ $[v,~ \omega]\gets$  \texttt{Wall\_Following($c_v, c_\omega,s_{WF} , r_{local}$)}}
		\Comment{See \ref{ap:wallfollowing}}
		\If{$x_{global}$ is in $list_{HP}$}
		\State{$state$ is "change\_local\_direction"}
		\EndIf
		\If{$M-line$ is hit and BA is closer to T} 
		\State $state \gets$ "rotate\_to\_target"
		\EndIf
		\ElsIf{$state$ is "rotate\_to\_target"}
		\State{$v \gets 0$} 
		\State{$\omega \gets c_\omega$}
		\If{Heading BA same as direction T}
		\State{$state \gets$ "forward"}
		\EndIf
		\ElsIf{$state$ is "change\_local\_direction"}
		\State{$v \gets 0$} 
		\State{$\omega \gets c_\omega$}
		\State{$s_{WF}=-1$}
		\If{BA has rotated 18$^o$}
		\State{$state \gets$ "wall\_following"}
		\EndIf	
		\EndIf	
		\Return {$v, \omega$}
		\EndFunction

	\end{algorithmic}
\end{algorithm}

\begin{algorithm}[H]
		\footnotesize
	\caption{The pseudo-code for the state-machine of Alg2. }\label{alg:alg2}
	\begin{algorithmic}
		
		\State Init:{ $state$ = "forward", $s_{WF}=1$, $list_{HP}$=[~]}
		\Require{$c_v$, $c_\omega, r_{local}$}
		\Function{Com}{}
		\If{$state$ is "forward" }

		\State{$v \gets c_v$} 
		\State{$\omega \gets 0$ } 
		\If{Obstacle is hit}
			\State{$s_{WF}=1$}
		\State {$d(H,T) \gets d(x_{global},T)for$}
		\State {$list_{HP}\gets[list_{HP}, x_{global}]$}
		\State {$state \gets$ "wall\_following"}
		\EndIf

		\ElsIf{$state$ is "wall\_following"}
		\State{ $[v,~ \omega]\gets$  \texttt{Wall\_Following($c_v, c_\omega,s_{WF} , r_{local}$)}}
		\Comment{See \ref{ap:wallfollowing}}
		\If{$x_{global}$ is in $list_{HP}$}
		\State{$state$ is "change\_local\_direction"}
		\EndIf
		\If{Way towards T is free and  $d(x_{global},T)<d(H,T)$} 
		\State $state \gets$ "rotate\_to\_target"
		\EndIf

		\ElsIf{$state$ is "rotate\_to\_target"}
		\State{$v \gets 0$} 
		\State{$\omega \gets c_\omega$}
		\If{Heading BA same as direction T}
		\State{$state \gets$ "forward"}
		\EndIf
		\ElsIf{$state$ is "change\_local\_direction"}
			\State{$v \gets 0$} 
			\State{$\omega \gets c_\omega$}
			\State{$s_{WF}=-1$}
			\If{BA has rotated 18$^o$}
			\State{$state \gets$ "wall\_following"}
			\EndIf
		\EndIf	
		\Return {$v, \omega$}
		\EndFunction

	\end{algorithmic}
\end{algorithm}
\renewcommand{\thesection}{B}
\section{Wall Following}

\subsection{Calculation Real Distance from Wall}\label{ap:realdistance}

\begin{figure}[H]
\centering
\includesvg[width=0.5\linewidth]{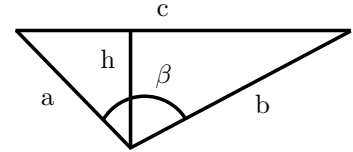}
	
	\caption{Visualization of the triangle configuration for the derivation.}
	\label{fig:rangetoheight}
\end{figure}

In Fig.~\ref{fig:rangetoheight}, the configurations of the solved triangle is solved, where we want to calculate $h$ (height triangle) with the triangle sides of $a$ and $b$ and the angle $\beta$. $c$ is the triangle side that will be unknown, so a formula will be derived that will only use $a$, $b$ and $\beta$.

The geometrical equations used to achieve the ranges are the triangle area formula:
\begin{equation}\label{eq:halvebase}
A = \frac{c \cdot h}{2}
\end{equation}
, the SAS triangle rule:
\begin{equation}\label{eq:SAS}
A = \frac{a\cdot b \cdot \sin \beta}{2}
\end{equation}
, the cosine-rule:
\begin{equation}\label{eq:cos}
c=\sqrt{a^2+b^2 - 2\cdot a \cdot b \cos \beta}
\end{equation}
with $A$ is the area of the triangle.

Substitute $A$ in Eq.~\ref{eq:SAS} for the right side of  Eq.~\ref{eq:halvebase}, and solve for $h$:

\begin{equation}\label{eq:halv2}
h=\frac{a\cdot b\cdot\sin\beta}{c}
\end{equation} 

Now substitute $c$ in Eq.~\ref{eq:halv2}, for the right side of Eq.~\ref{eq:cos}, which results in the following equation:

\begin{equation}
h=\frac{a\cdot b\sin\beta}{\sqrt{a^2+b^2 - 2\cdot a \cdot b \cos \beta}}
\end{equation} \label{eq:range2height}

\subsection{Pseudo Code Wall Following} \label{ap:wallfollowing}

The procedure of the wall-following behavior is listed in this appendix in Algorithm~\ref{alg:wallfollowing} and~\ref{alg:wallfollowingaligning}. T$s_{WF}$ is a variable that determines if the wall-following is right- ($s_{WF}$=1) or left-sided ($s_{WF}$=-1), $d(x,O_\perp)$ is the current distance to the robot calculated perpendicular from the wall and $d_{ref}$ is the preferred distance from the wall in meters and $t_d$ is the threshold to determine if the robot near $d_{ref}$. $r_s$ and $r_f$ are the side and range sensor's measurement in meters and $\beta$ is the angle between them. If $s_{WF}$=1, then $r_s$ is the right range sensor and if $s_{WF}$=-1, then $r_s$ is the left range sensor. $v$ is the control output for the forward velocity of the robot and $c_v$ is the fixed velocity constant. $\omega$ is the control output for the heading of the robot in rad/s and $c_\omega$ is a fixed rate constant, to control the speed of the robot's turns.
\setcounter{algorithm}{0}
\begin{algorithm}[H]
		\footnotesize
	\caption{The procedure of the wall-following behavior.}\label{alg:wallfollowing}
	\begin{algorithmic}

\State Init: $state$ = "rotate\_to\_align\_wall"
\Require{ $s_{WF},~ d(x,O_\perp),~ d_{ref},~ r_s,~ r_f,~ c_v, c_\omega,~ \beta$}
\State $\beta$ = $60\deg$
\Function{Wall\_Following}{}
\If{$state$ is "rotate\_to\_align\_wall" }
    \State{$v \gets 0$}
    \State{$\omega \gets -1 \cdot s_{WF} \cdot c_\omega$ } \Comment{Turn away from the wall}

	\If{$r_s\approx r_f\cdot\cos(\beta)$}
		\State $state \gets$ "wall\_following\_and\_aligning"
	\EndIf
	\If{$r_f$ = OR}
		\State $state \gets$ "rotate\_around\_corner"
	\EndIf
\ElsIf{$s_{WF}$ is "wall\_following\_and\_aligning"}
	\State{$v\gets c_v$}
    \State $\omega\gets$  \texttt{Wall\_Following\_and\_Aligning()}
    \Comment{See \ref{alg:wallfollowingaligning}}
	\If{$d(x,O_{min}) < d_{ref}$}
		\State $state \gets$ "rotate\_to\_align\_wall"
	\EndIf
	\If{$r_f$ is OR}
		\State $state \gets$ "rotate\_around\_corner"
	\EndIf
\ElsIf{$state$ is "rotate\_around\_corner"}
	  \State{$v \gets c_v$}
	  \State{$\omega \gets s_{WF}\cdot v / d_{ref}$} 
	      \Comment{Wide turn, radius = $d_{ref}$ }
	\If{$r_s\approx r_f\cdot\cos(\beta)$}
	\State $state \gets$ "wall\_following\_and\_aligning"
	\EndIf
	\If{$d(x,O_{min}) < d_{ref}$}
	\State $state \gets$ "rotate\_to\_align\_wall"
	\EndIf
\EndIf

		\Return $\omega$
\EndFunction

	\end{algorithmic}
\end{algorithm}

\begin{algorithm}[H]
		\footnotesize
	\caption{The procedure of  keeping the heading of the FootBot aligned with the wall during wall following. }\label{alg:wallfollowingaligning}
	\begin{algorithmic}
		
		\Require{ $s_{WF},~ d(x,O_\perp),~ d_{ref},~ r_s,~ r_f,~ c_w,~ \beta$}
		
		\Function{Wall\_Following\_And\_Aligning}{}		
		\If{$|d_{ref}-d(x,O_\perp)|>-t_d$ )} \Comment{If too far from $d_{ref}$}
		\If {$d_{ref}-d(x,O_\perp)>t_d$} \Comment{If too far from wall}

		\State $\omega =  s_{WF} \cdot c_\omega$   \Comment{Turn towards the wall} 
		
		\Else\Comment{If too close to wall}
		
		\State $\omega = - s_{WF} \cdot c_\omega$ \Comment{Turn from the wall }
		\EndIf
		
		\ElsIf {$|d_{ref}-d(x,O_\perp)|<t_d$} \Comment{If close to $d_{ref}$}
		\If{$r_s > r_f \cdot \cos \beta$ } \Comment{Fine tune alignment}
		\State $\omega =  s_{WF} \cdot c_\omega$ \Comment{Turn towards the wall}
		\Else
		\State $\omega =  -s_{WF} \cdot c_\omega$ \Comment{Turn from the wall}
		\EndIf
		\Else \Comment{Do not adjust the turn}
		\State $\omega = 0$
		\EndIf
		
		\Return $\omega$
		\EndFunction
	\end{algorithmic}
\end{algorithm}

\renewcommand{\thesection}{C}
\section{Statistical Tests}

\subsection{Bootstrapping Bug Algorithms}\label{ap:bootbas}

In Fig.~\ref{fig:resultsbugnonoise}, the resulting performance values per bug algorithm was shown. Here, both the success rate and the trajectory length are subjected to a bootstrapping test, to evaluate whether the bug algorithms belong to the same distribution (null-hypothesis). Table~\ref{table:bootstraptrajectory} contains the boostrapping tests from the data presented in Fig.~\ref{fig:resultsbugnonoise}(a) and Table~\ref{table:bootstrapsuccess} for Fig.~\ref{fig:resultsbugnonoise}(b).

\begin{table}[H]
\centering
\footnotesize
\begin{tabular}{|c|c|c|c|c|c|}
\hline
 & Com & Com1 & Bug2 & Alg1 & Alg2 \\
\hline
Com & 1 & 1 & 0 & 0 & 1 \\
\hline
Com1 & 1 & 1 & 0 & 0 & 1 \\
\hline
Bug2 & 0 & 0 & 1 & 1 & 0 \\
\hline
Alg1 & 0 & 0 & 1 & 1 & 0 \\
\hline
Alg2 & 1 & 1 & 0 & 0 & 1 \\
\hline
\end{tabular}
\caption{Bootstrapping results on the trajectory length of the evaluated bug algorithms with a sample size 10000. The value "1" means that the null-hypothesis (the evaluated data comes from the same distribution) holds, while "0" means it is rejected.}
\label{table:bootstraptrajectory}
\end{table}

\hspace{-0.5cm}\begin{table}[H]
\centering
\footnotesize
\begin{tabular}{|c|c|c|c|c|c|}
\hline
 & Com & Com1 & Bug2 & Alg1 & Alg2 \\
\hline
Com & 1 & 0 & 0 & 0 & 0 \\
\hline
Com1 & 0 & 1 & 1 & 1 & 0 \\
\hline
Bug2 & 0 & 1 & 1 & 1 & 0 \\
\hline
Alg1 & 0 & 1 & 1 & 1 & 0 \\
\hline
Alg2 & 0 & 0 & 0 & 0 & 1 \\
\hline
\end{tabular}
\caption{Bootstrapping results on the success rate of the evaluated bug algorithms with a sample size 10000. The value "1" means that the null-hypothesis (the evaluated data comes from the same distribution) holds, while "0" means it is rejected.}
\label{table:bootstrapsuccess}
\end{table}

\subsection{Correlation Analysis Odometry Noise}\label{ap:corrodometry}

In order to evaluate whether an relationship exists between the increasing odometry noise and the degeneration of the performances of the bug algorithms, the data presented in Fig.~\ref{fig:resultsbugwithnoise} are subjected to regression analysis. Table~\ref{table:logregodometry} contains the logistic regression analysis with a R2 value, from the trajectory length data presented in Fig.~\ref{fig:resultsbugwithnoise}(a) and Table~\ref{table:linregodometry} contains the logistic regression analysis with a pseudo-R2 value, from the success rate data presented in Fig.~\ref{fig:resultsbugwithnoise}(b).

\begin{table}[H]
\centering
\footnotesize
\begin{tabular}{|c|c|c|c|c|c|}
\hline
 & Com & Com1 & Bug2 & Alg1 & Alg2 \\
\hline
Slope & 8.081 & 13.642 & 13.561 & 11.857 & 12.523 \\
\hline
Intercept & 2.752 & 2.724 & 3.152 & 3.522 & 2.654 \\
\hline
R2 & 0.076 & 0.189 & 0.217 & 0.173 & 0.161 \\
\hline
\end{tabular}
\caption{Linear regression evaluation of the trajectory lengths against the measurement noise, including the intercept, slope and R2 value per bug algorithm.}
\label{table:linregodometry}
\end{table}

\begin{table}[H]
\centering
\footnotesize
\begin{tabular}{|c|c|c|c|c|c|}
\hline
 & Com & Com1 & Bug2 & Alg1 & Alg2 \\
\hline
Slope & -1.240 & -3.100 & -3.860 & -3.480 & -3.110 \\
\hline
Intercept & 0.587 & 0.800 & 0.779 & 0.706 & 0.847 \\
\hline
R2 & 0.035 & 0.189 & 0.343 & 0.323 & 0.198 \\
\hline
\end{tabular}
\caption{Logistic regression evaluation of the success rate against the measurement noise, including the intercept, slope and (psuedo) R2 value per bug algorithm.}
\label{table:logregodometry}
\end{table}

\subsection{Correlation Analysis Recognition Failures}\label{ap:corrFNFP}

In order to evaluate whether an relationship exists between the increasing failing recognition rate and the degeneration of the performances of the bug algorithms Alg1 and Alg2, the data presented in Fig.~\ref{fig:resultsbugwithFP} are subjected to regression analysis. Table~\ref{table:logregFP} contains the logistic regression analysis with a R2 value, from the trajectory length data presented in Fig.~\ref{fig:resultsbugwithFP}(a) and Table~\ref{table:linregFP} contains the logistic regression analysis with a pseudo-R2 value, from the success rate data presented in Fig.~\ref{fig:resultsbugwithFP}(b).
Table~\ref{table:logregFN} contains the logistic regression analysis with a R2 value, from the trajectory length data presented in Fig.~\ref{fig:resultsbugwithFP}(c) and Table~\ref{table:linregFN} contains the logistic regression analysis with a pseudo-R2 value, from the success rate data presented in Fig.~\ref{fig:resultsbugwithFP}(d).

\begin{table}[H]
\centering
\footnotesize
\begin{tabular}{|c|c|c|}
\hline
 & Alg1 & Alg2 \\
\hline
Slope & 0.5472 & 0.1722 \\
\hline
Intercept & 2.4302 & 1.8843 \\
\hline
R2 & 0.0112 & 0.0020 \\
\hline
\end{tabular}
\caption{Linear regression evaluation of the trajectory lengths against the False Positive recognition rate, including the intercept, slope and R2 value per bug algorithm.}
\label{table:linregFP}
\end{table}

\begin{table}[H]
\centering
\footnotesize
\begin{tabular}{|c|c|c|}
\hline
 & Alg1 & Alg2 \\
\hline
Slope & -0.1443 & -0.0014 \\
\hline
Intercept & 0.9105 & 0.9418 \\
\hline
R2 & 0.4652 & 0.7773 \\
\hline
\end{tabular}
\caption{Logistic regression evaluation of the success rate against the False Positive recognition rate, including the intercept, slope and (psuedo) R2 value per bug algorithm.}
\label{table:logregFP}
\end{table}

\begin{table}[H]
\centering
\footnotesize
\begin{tabular}{|c|c|c|}
\hline
 & Alg1 & Alg2 \\
\hline
Slope & 0.1873 & 1.0584 \\
\hline
Intercept & 3.2478 & 2.2733 \\
\hline
R2 & 0.0000 & 0.0006 \\
\hline
\end{tabular}
\caption{Linear regression evaluation of the trajectory lengths against the False Negative recognition rate, including the intercept, slope and R2 value per bug algorithm.}
\label{table:linregFN}
\end{table}

\begin{table}[H]
\centering
\footnotesize
\begin{tabular}{|c|c|c|}
\hline
 & Alg1 & Alg2 \\
\hline
Slope & 0.0577 & 0.1010 \\
\hline
Intercept & 0.8018 & 0.9192 \\
\hline
R2 & 0.3668 & 0.7171 \\
\hline
\end{tabular}
\caption{Logistic regression evaluation of the success rate against the False Negative recognition rate, including the intercept, slope and (psuedo) R2 value per bug algorithm.}
\label{table:logregFN}
\end{table}

\subsection{Correlation Analysis Distance Sensor Noise}\label{ap:corrrange}

In order to evaluate whether an relationship exists between the increasing distance measurement noise and the degeneration of the performances of the bug algorithms Alg1 and Alg2, the data presented in Fig.~\ref{fig:resultsbugwithrangenoise} are subjected to regression analysis. Table~\ref{table:logregRange} contains the logistic regression analysis with a R2 value, from the trajectory length data presented in Fig.~\ref{fig:resultsbugwithrangenoise}(a) and Table~\ref{table:linregRange} contains the logistic regression analysis with a pseudo-R2 value, from the success rate data presented in Fig.~\ref{fig:resultsbugwithrangenoise}(b).
\begin{table}[H]
\centering
\footnotesize
\begin{tabular}{|c|c|c|}
\hline
 & Com1 & Alg2 \\
\hline
Slope & 0.0501 & -0.0075 \\
\hline
Intercept & 2.5783 & 2.4204 \\
\hline
R2 & 0.0019 & 0.0001 \\
\hline
\end{tabular}
\caption{Linear regression evaluation of the trajectory lengths against the distance measurement noise, including the intercept, slope and R2 value per bug algorithm.}
\label{table:linregRange}
\end{table}

\begin{table}[H]
\centering
\footnotesize
\begin{tabular}{|c|c|c|}
\hline
 & Com1 & Alg2 \\
\hline
Slope & -0.0557 & -0.0225 \\
\hline
Intercept & 0.8682 & 0.9283 \\
\hline
R2 & 0.2412 & 0.5583 \\
\hline
\end{tabular}
\caption{Logistic regression evaluation of the success rate against the distance measurement noise, including the intercept, slope and (psuedo) R2 value per bug algorithm.}
\label{table:logregRange}
\end{table}

\end{document}